%% file: main.tex
\documentclass[10pt,journal,compsoc]{IEEEtran}       

\usepackage{amsmath}
\usepackage{amsfonts}
\usepackage{mathtools}
\usepackage{amssymb}

\DeclarePairedDelimiter\floor{\lfloor}{\rfloor}
\usepackage[utf8]{inputenc}
\usepackage[numbers,sort]{natbib}
\usepackage{todonotes}
\usepackage{url}
\usepackage{booktabs}

\usepackage{xspace}
\newcommand{\mathsymbol}[2]{\newcommand{#1}{\ensuremath{\mathit{#2}}\xspace}}
\newcommand{\textmacro}[2]{\mathsymbol{#1}{\text{#2}}}
\mathsymbol{\hyperband}{\text{\sc Hyperband}}
\textmacro{\mlplan}{ML-Plan}
\textmacro{\autoweka}{Auto-WEKA}
\textmacro{\autosklearn}{auto-sklearn}
\textmacro{\tpot}{TPOT}
\textmacro{\recipe}{RECIPE}

\mathsymbol{\dataspace}{D}
\mathsymbol{\dataset}{d}
\mathsymbol{\datasethalf}{\dataset_\mathit{half}}
\mathsymbol{\datasetfull}{\dataset_\mathit{full}}
\mathsymbol{\instancespace}{\mathcal{X}}
\mathsymbol{\labelspace}{\mathcal{Y}}
\mathsymbol{\hypospace}{H}
\mathsymbol{\learner}{a}
\mathsymbol{\learnerset}{A}
\mathsymbol{\learnerspace}{\mathcal{A}}
\mathsymbol{\risk}{\mathcal{R}}
\mathsymbol{\selector}{\sigma}
\mathsymbol{\valid}{\nu}
\mathsymbol{\validsmart}{\tilde\nu}
\mathsymbol{\void}{\bot}
\mathsymbol{\lc}{p}
\mathsymbol{\schedule}{S}
\mathsymbol{\anchor}{s}
\mathsymbol{\runtimecv}{\tau_{cv}}
\mathsymbol{\runtimelccv}{\tau_{lccv}}
\mathsymbol{\minexp}{\rho}
\textmacro{\automl}{AutoML}

\mathsymbol{\prunethreshold}{r}
\mathsymbol{\maxciwidth}{\varepsilon}
\mathsymbol{\toleranceparam}{\gamma}
\mathsymbol{\utilitythreshold}{\beta}
\mathsymbol{\restrictionball}{q}
\mathsymbol{\empmean}{\hat \mu}

\newtheorem{theorem}{Theorem}

\usepackage[ruled,vlined,linesnumbered]{algorithm2e}

\begin{document}


\title{Fast and Informative Model Selection using Learning Curve Cross-Validation}
\author{Felix Mohr, Jan N. van Rijn}

\IEEEtitleabstractindextext{%
\begin{abstract}
    Common cross-validation (CV) methods like k-fold cross-validation or Monte-Carlo cross-validation estimate the predictive performance of a learner by repeatedly training it on a large portion of the given data and testing on the remaining data.
    These techniques have two major drawbacks.
    First, they can be unnecessarily slow on large datasets.
    Second, beyond an estimation of the final performance, they give almost no insights into the learning process of the validated algorithm.
    In this paper, we present a new approach for validation based on learning curves (LCCV).
    Instead of creating train-test splits with a large portion of training data, LCCV iteratively increases the number of instances used for training.
    In the context of model selection, it discards models that are very unlikely to become competitive.
    We run a large scale experiment on the 67 datasets from the \automl benchmark and empirically show that in over 90\% of the cases using LCCV leads to similar performance (at most 1.5\% difference) as using 5/10-fold CV.
    However, it yields substantial runtime reductions of over 20\% on average.
    Additionally, it provides important insights, which for example allow assessing the benefits of acquiring more data.
    These results are orthogonal to other advances in the field of \automl.
\end{abstract}
}

{\maketitle}

\section{Introduction}
Model selection is the process of picking from a given set of machine learning models the one that is believed to exhibit the best predictive performance.
An important sub-task of model selection is to estimate the performance of each candidate model, which is done by so-called \emph{validation methods}.
For example, tools for automated machine learning (\automl) make excessive use of validation techniques to assess the performance of machine learning pipelines \cite{autoweka,feurer2015efficient,olson2016tpot,mohr2018ml}.
The validation procedure can often be considered just an interchangeable function that provides some estimate for the quality of a model.
Typical choices to implement such a validation function are k-fold cross-validation (kCV), and Monte-Carlo Cross-Validation (MCCV).

Those validation methods are often unnecessarily slow on large datasets.
They use a large portion of the data for learning a model, and the size of this portion is a constant that is independent of the data over which is evaluated.
For example, kCV and MCCV typically use between 70\% and 90\% of the data for training.
While large training portions are \emph{sometimes} necessary to estimate the generalization performance of a learner, in many cases the number of data points required to build the best possible model of a class is much lower.
Then, large train folds unnecessarily slow down the validation process, sometimes substantially.
For example, the error rate of a linear SVM on the {\sc numerai28.6} dataset is 48\% when training with $4\,000$ instances but also when training with $90\,000$ instances.
However, in the latter case, the evaluation is almost 50 times slower. 

Additionally, vanilla validation procedures provide little additional information to the data owner.
Data owners usually need to ask themselves the question which of the conventional options would improve the quality of their current model: (i)~gather more data points, (ii)~gather more attributes, or (iii)~apply a wider range of models. 
Common validation techniques only provide information for (iii).

In this paper, we propose a cross-validation approach based on the notion of learning curves.
Learning curves express the predictive performance of the models produced by a learner for different numbers of instances available for learning.
The validation of a learner is now done as follows.
Instead of evaluating it just for one number of training instances (say 80\% of the original dataset size), it is evaluated, in increasing order, at different so-called \emph{anchor points} of training fold sizes, e.g. for 64, 128, 256, 512, ... instances.
At each anchor point, several evaluations are conducted until a stability criterion is met.
We dub this approach learning-curve-based cross-validation (LCCV).
Other authors also utilized learning curves~\cite{baker2017accelerating,domhan2015speeding,klein2017learning,leite2010active,rijn2015fast} or sub-sampling~\cite{jamieson2016non,lihyperband2017,petrak2000fast,provost1999efficient,sabharwal2016selecting} to speed-up model selection by discarding unpromising candidates early.
The main contribution of LCCV is that it
\emph{combines} empirical learning curves with an \emph{convexity assumption} on learning curves to early prune candidates only if they are unlikely to be an optimal solution.
LCCV does not rely on a learning curve model like an inverse power law (IPL) for this, even though such a model is used for other decisions in the evaluation process.

Our work focuses on the scenario in which learners are trained in a non-incremental way.
At first sight, LCCV may seem a natural fit when the portfolio from which a model is to be selected contains incremental learners.
Learning curves essentially come for free for such learners since we can build them during the training process over the different iterations.
This seems like an advantage over observation-based learning curves, which may require training from scratch at each anchor and hence more time to be computed.
However, there are some issues with incremental learning curves on both the practical and theoretical side.
First, even though incremental versions have been proposed for many learners, these are often not realized in widely used implementations like WEKA \cite{hall2009weka} or scikit-learn \cite{pedregosa2011scikit} and hence not available for data scientists using those libraries.
Second, incremental learning curves are of a different \emph{type} than the observation-based learning curves~\cite{domhan2015speeding}.
There is empirical evidence that suggests that observation-based learning curves are convex (see Appendix~\ref{app:convexity} in the supplement), whereas this is less clear for the iteration-based learning curves~\cite{viering2021theshape}.
Since our work relies on the convexity assumption, we focus on non-incremental training.


In such scenarios, LCCV is an interesting alternative to classical validation techniques.
At first sight, it seems to suffer from the drawback that it builds models at different anchors from scratch and hence to be even \emph{slower} than classical validation techniques.
However, we show in Sec. \ref{sec:evaluation:randomsearch} that this is rarely the case.
In fact, we consider that LCCV has three advantages over those techniques specifically in the context of AutoML
\cite{feurer2015efficient,olson2016tpot,mohr2021towards}:
\begin{enumerate}
    \item When used for model selection without or with moderate timeouts, LCCV exploits a convexity assumption to early discard candidates that are unlikely to become competitive.
    Thereby, it chooses in almost all cases a model that is equally good as the one chosen by common cross-validation but usually in substantially less time.

    \item When used for model selection with tight timeouts, LCCV is more reliable in its ability to produce \emph{any} result for a candidate.
    Even with a very limited time budget, LCCV yields at least an \emph{estimate} of the reference model performance even if not with the same confidence as if no timeout is applied.
    In particular, LCCV can return estimates in situations where there is \emph{no} score obtained using common cross-validation techniques.
    
    \item Instead of a single-point observation, LCCV gives insights about the learning \emph{behaviour} of a candidate.
    This is very useful to answer the question of whether more data would help to obtain better results; information that cannot at all be provided by any of the above validation techniques.
    Note that this information cannot generally be obtained via standard cross-validation, e.g., by afterwards computing the learning curve of the learner that was best in cross-validation on, say, 80\% of the training data.
    This is because there is no guarantee that this learner (or even only any of the $k$ best learners) would also be the best if more data were available; this is precisely something that can be revealed with LCCV since it computes the learning curves of \emph{all} learners.
    
\end{enumerate}

We sustain these claims empirically by comparing LCCV against kCV, the arguably most popular validation technique.
First, we show that LCCV is superior to both 5CV and 10CV in terms of runtime in most situations.
Our experiments compare the runtimes of a random search using LCCV or kCV for validation and confirm that LCCV most often yields similar results as 5CV and 10CV while reducing runtime on average by 20\%, in an experiment where the task is to select a learner from a portfolio of 200 machine learning pipelines.
In absolute numbers, in this experiment the runtime of a single model selection process is be reduced, on average, by 1.5h compared to 5CV and even 2.75h compared to 10CV.
In addition, the experiments show that the learning curves obtained using LCCV can be directly used to assert whether or not more data could improve the overall result of the model selection process.

\section{Problem Statement}
\label{sec:problem}
Our focus is on supervised machine learning.
For an \emph{instance space} \instancespace and a \emph{label space} \labelspace, a \emph{dataset} $\dataset \subset \{(x,y)~|~x\in \instancespace, y \in \labelspace\}$ is a \emph{finite} relation between the two spaces.
We denote as \dataspace the set of all possible datasets.
A \emph{learning algorithm} is a function $\learner: \dataspace \rightarrow \hypospace$, where $\hypospace = \{h~|~h: \instancespace \rightarrow \labelspace\}$ is the space of hypotheses.
The performance of a hypothesis is its \emph{out-of-sample error}
\begin{equation}\label{eq:risk}
\risk(h) = \int\limits_{\instancespace,\labelspace}{loss(y, h(x))} \, d \mathbb{P}(x,y),
\end{equation}
and the performance of a learner \learner when trained with some data \dataset is $\risk(\learner(\dataset))$.
Here, $loss(y,h(x)) \in \mathbb{R}$ is the penalty for predicting $h(x)$ for instance $x \in \instancespace$ when the true label is $y \in \labelspace$, and $\mathbb{P}$ is a joint probability measure on $\instancespace \times \labelspace$ from which the available dataset \dataset has been generated.

Since the performance in Eq. (\ref{eq:risk}) cannot be evaluated, an estimate is needed.
The only known feasible way to do this is by validation, i.e. by splitting the data \dataset into $\dataset_{train}$ and $\dataset_{validate}$, to run the learning algorithm (learner) only with $\dataset_{train}$, and to use $\dataset_{validate}$ to \emph{estimate} the risk.
Formally, we then have a \emph{validation} function
\begin{equation}
    \valid: \dataspace \times \learnerset \rightarrow \mathbb{R},
\end{equation}
such that $\valid(\dataset, \learner)$ is an estimate of the out-of-sample error of the hypothesis produced by the learner \learner based on the available data \dataset.
Typically, \valid will not only create one but several splits, train the learner several times and aggregate the prediction results into an overall estimate.

Several learning algorithms use such a validation function as a \emph{sub-routine} to select a model among a set of candidates it has produced.
For example, \automl tools generate and evaluate sequences of models.
Formally, such learners produce a sequence $\learnerset = (\learner_1, \learner_2, ...)$ of learners, each of which is validated with \valid, and eventually the (hypothesis produced by the) learner with the lowest empirical risk is returned in the hope that this also minimizes the out-of-sample error in Eq. (\ref{eq:risk}).

We follow the idea that the score of a non-competitive candidate does not even need to be computed exactly~\cite{domhan2015speeding}.
In other words, we do not even need to know the concrete score $\valid(\dataset, \learner)$ if $\mathbb{E}[\risk(\learner(\dataset_{train}))] > \prunethreshold$, where the expectation is taken with respect to the train data sets $\dataset_{train}$ with the same size as being used by \valid and where \prunethreshold is the currently best observed performance.
This scenario motivates an \emph{informed} validation function
\begin{equation}
    \validsmart: \dataspace \times \learnerset \times \mathbb{R} \rightarrow \mathbb{R} \cup \{\void\},
\end{equation}
which takes as an additional argument the reference score and is allowed to return a result \void to indicate that the candidate can be discarded.
Even if there is some tentative performance estimate of the candidate, \validsmart should not return that estimate as it is not necessarily comparable to previous results and could be potentially misleading the search.

While the exact validation result of non-competitive candidates can be important for some search algorithms, there are many situations where this is not the case.
Indeed, several search strategies \emph{steer} the search process based on the outcome of earlier validations (e.g., Bayesian optimization, evolutionary algorithms, or tree search). 
So returning an uninformative value \void that leads to discarding of the candidate might have undesirable side effects.
On the other hand, existing state of the art approaches using Bayesian optimization adopting such a technique have apparently not been affected by it \cite{domhan2015speeding}, and other (tree-based) approaches are even invariant to this issue by concept \cite{mohr2018ml}.
As such, we do not focus on issues possibly caused by returning a void value \void in this paper.

The goal is to find an implementation of \validsmart that contributes to some external objectives such as runtime minimization or generating interesting insight while providing a reasonably good estimate of the out-of-sample error.
The quality of \validsmart with respect to this constraint is operationalized by a base validation function \valid considered as ground truth so that
\begin{equation}
\label{eq:constraintonsmartvalidation}
\validsmart(\dataset, \learner, \prunethreshold) =
\begin{cases}
    \bot & \text{only if }\valid(\dataset, \learner) > \prunethreshold\\
    \in B_\restrictionball(\valid(\dataset, \learner)) & \text{else}
\end{cases},
\end{equation}
where $B_\restrictionball$ is some $\restrictionball$-neighbourhood around the actual validation score $\valid(\dataset, \learner)$.
In practice, \valid can be realized through an MCCV with a high number of repeats.

Note that the proposed \validsmart is a simplification of a normal validation method $v$. However, utilizing this simplification, we will show that we can obtain better runtimes, which is the main contribution of this work. 
In Section~\ref{sec:lccv}, we will present an algorithm that utilizes learning curves to achieve this, and as a side product gives additional insights into the learning procedure. 

\section{Related Work}
Model selection is at the heart of many data science approaches. 
When provided with a new dataset, a data scientist is confronted with the question of which model to apply to this. 
This problem is typically abstracted as the \emph{combined algorithm selection and hyperparameter optimization problem}~\cite{autoweka}. 
To properly deploy model selection methods, there are three important components: 
\begin{enumerate}
  \item A \emph{configuration space}, which specifies the set of algorithms to be considered and, perhaps, their hyperparameter domains
  \item A \emph{search procedure}, which determines the order in which these algorithms are considered
  \item An \emph{evaluation mechanism}, which assesses the quality of a certain algorithm
\end{enumerate}
Most research addresses the question how to efficiently search the configuration space, leading to a wide range of methods such as random search~\cite{bergstra2012random}, Bayesian optimization~\cite{bergstra2011algorithms,hutter2011sequential,snoek2012practical}, evolutionary optimization~\cite{loshchilov2016cma}, meta-learning~\cite{brazdil2008meta,pinto2016towards} and planning-based methods~\cite{mohr2018ml}.

Our work aims to make the evaluation mechanism faster, while at the same time not compromising the performance of the algorithm selection procedure. 
As such, it can be applied orthogonal to the many advances made on the components of the configuration space and the search procedure. 

The typical methods used as evaluation mechanisms are using classical methods such as a holdout set, cross-validation, leave-one-out cross-validation, and bootstrapping. 
This can be sped up by applying racing methods, i.e., to stop evaluation of a given model once a statistical test lets the possibility of improving over the best seen so far appear unlikely.
Some notable methods that make use of this are ROAR~\cite{hutter2011sequential} and iRace~\cite{lopez2016irace}.
The authors of~\cite{feurer2018towards} focus on setting the hyperparameters of AutoML methods, among others selecting the evaluation mechanism. 
Additionally, model selection methods are accelerated by considering only subsets of the data.
By sampling various subsets of the dataset of increasing size, one can construct a learning curve.
While these methods at their core have remained largely unchanged, there are two directions of research building upon this basis: (i) model-free learning curves, and (ii) learning curve prediction models. 

{\bf Model-free learning curves:}
The simplest thing one could imagine is training and evaluating a model based upon a small fraction of the data~\cite{petrak2000fast}.
The authors of~\cite{provost1999efficient} propose progressive sampling methods using batches of increasing sample sizes (which we also leverage in our work) and propose mechanisms for detecting whether a given algorithm has already converged. 
However, the proposed convergence detection does not take into account randomness from selecting a given set of sub-samples, making the method fast but at risk of terminating early. 

The authors of~\cite{wang2021autods} introduce FLAML, which uses multiple sample sizes, by adding a sample size hyperparameter to the configuration space. The reasoning behind this design choice is to allow low complexity learners to be tested on small amounts of data, and high complexity learners to be tested on larger amounts of data. FLAML starts with evaluating configurations on low sample sizes and gradually extends this to larger sample sizes. 

Successive halving addresses the model selection problem as a bandit-based problem, that can be solved by progressive sampling ~\cite{jamieson2016non}.
All models are evaluated on a small number of instances, and the best models get a progressively increasing number of instances.
While this method yields good performance, it does not take into account the development of learning curves, e.g., some learners might be slow starters, and will only perform well when supplied with a large number of instances.
For example, the extra tree classifier~\cite{geurts2006extremely} is the best algorithm on the {\sc PhishingWebsites} dataset when training with all ($10\,000$) instances but the \emph{worst} when using $1\,000$ or fewer instances; it will be discarded by successive halving (based on all scikit-learn algorithms with default hyperparameters).
Hyperband aims to address this by building a loop around successive halving, allowing learners to start at various budgets~\cite{lihyperband2017}. 
While the aforementioned methods all work well in practice, these are all greedy in the fact that they might disregard a certain algorithm too fast, leading to fast model selection but sometimes sub-optimal performances.

Another important aspect is that the knowledge of the whole portfolio plays a key role in successive halving and Hyperband.
Other than our approach, which is a \emph{validation} algorithm and does not have knowledge about the portfolio to be evaluated (and makes no global decisions on budgets), successive halving assumes that the portfolio is already defined and given, and Hyperband provides such portfolio definitions.
In contrast, our approach will just receive a sequence of candidates for validation, and this gives more flexibility to the approaches that want to use it.

The authors of~\cite{sabharwal2016selecting} define the cost-sensitive training data allocation problem, which is related to the problem that we aim to address. 
They also introduce an algorithm that solves this problem, i.e., Data Allocation using Upper Bounds. Given a set of configurations, it first runs all configurations on two subsamples of the dataset, effectively building the initial segment of the learning curve. Based on this initial segment per configuration, it determines for each an optimistic upper bound.
After that, it goes into the following loop:
It runs the most promising configuration on a larger sample size and updates the performance bound. It reevaluates which configuration has the most potential at that moment, and continues with assigning more budget to the most promising configuration until one configuration has been run on the entire dataset.
Similar to Hyperband and in contrast to LCCV, it needs to know the entire portfolio of learners at the start. 

{\bf Learning curve prediction models: }
A model can be trained based on the learning curve, predicting how it will evolve. 
The authors of~\cite{domhan2015speeding} propose a set of parametric formulas to which can be fitted so that they model the learning curve of a learner.
They employ the method specifically to neural networks, and the learning curve is constructed based on epochs, rather than instances. 
This allows for more information about earlier stages of the learning curve without the need to invest additional runtime. 
By selecting the best fitting parametric model, they can predict the performance of a certain model for a hypothetical increased number of instances. 
The authors of~\cite{klein2017learning} build upon this work by incorporating the concept of Bayesian neural networks.

When having a set of historic learning curves at disposition, one can efficiently relate a given learning curve to an earlier seen learning curve, and use that to make predictions about the learning curve at hand.
The authors of~\cite{leite2010active} employ a k-NN-based model based on learning curves for a certain dataset to determine which datasets are similar to the dataset at hand. 
This approach was later extended by the authors of~\cite{rijn2015fast}, to also select algorithms fast. 

Similar to our approach, the authors of~\cite{baker2017accelerating} proposed a model-based version of Hyperband.
They train a model to predict, based on the performance of the last sample, whether the current model can still improve upon the best-seen model so far. 
Like all model-based learning curve methods, this requires vast amounts of meta-data, to train the meta-model.
And like the aforementioned model-free approaches, these model-based approaches are at risk of terminating a good algorithm too early. 
In contrast to these methods, our method is model-free and always selects the optimal algorithm based on a small set of reasonable assumptions. 

\section{Learning-Curve-Based Cross-Validation}\label{sec:lccv}
Learning-curve-based cross-validation (LCCV) estimates the performance of a learner based on empirical learning curves.
LCCV evaluates the learner on ascendingly ordered so-called \emph{anchor (points)} $S = (s_1,..,s_T)$, which are training set sizes.
Several such evaluations are made on each anchor.
The idea is to early recognize whether or not a candidate can be competitive.
Sec. \ref{sec:lccv:cancellation} explains how we use the convexity of the learning curve to decide on this question, and Sec. \ref{sec:lccv:convexitycompatibleobservationsets} details how we make sure that this process only discards candidates that are sub-optimal with high probability.
Sec. \ref{sec:lccv:skippingevaluations} explains how we use inverse power law (IPL) estimates to skip anchor points and directly jump to $s_T$ for presumably competitive candidates.
The algorithm is formalized in Sec. \ref{sec:lccv:algorithm}.
That section also outlines that, even if the early stopping or skipping mechanisms fail, the runtime of LCCV is in the worst case twice as high as the one of a kCV.

\subsection{Aborting a Validation}
\label{sec:lccv:cancellation}
The idea of LCCV is that we can discard a candidate early with high confidence if it is unlikely to be competitive. 
For this, we assume that its learning curve is \emph{convex}.
It is known that not all learning curves are convex, but we conducted an extensive empirical study (see Appendix~\ref{app:convexity} in the supplementary material) justifying this assumption for the large majority of (the analyzed but presumably also other) cases.
Convexity is important because it implies that the \emph{slope} of the learning curve is an \emph{increasing} function that, for increasing training data sizes, approximates 0 from below (and sometimes even exceeds it).
Recall that we assume error curves in this paper; for accuracy curves or F1 curves etc., the curve slopes would be decreasing of course.
The convexity implies that we can take the observations of the last two anchors of our \emph{empirical} learning curve, compute the slope, and extrapolate the learning curve with this slope.
From this, we obtain a \emph{bound} (not an estimate opposed to \cite{domhan2015speeding}) for the performance at the full data size and can prune if this bound will be worse than threshold value \prunethreshold. This is in line with the work of~\cite{sabharwal2016selecting}.

More formally, denote the learning curve as a function $\lc(\anchor)$, where \lc is the \emph{true average} performance of the learner when trained with a train set of size \anchor.
Convexity of \lc implies that $\lc'(\anchor) \leq \lc'(\anchor')$ if $\anchor' > \anchor$, where $\lc'$ is the derivative of \lc.
Of course, as a learning curve is a sequence, it is not derivable in the strict sense, so we rather sloppily refer to its slope $\lc'$ as the slope of the line that connects a point with its predecessor, i.e., $\lc'(\anchor) := \lc(\anchor) - \lc(\anchor - 1)$ for all $\anchor \in \mathbb{N}_+$; note that based on the convexity assumption $\lc' \leq 0$.
If we find, for some \anchor, that $\lc(\anchor) + \lc'(\anchor)\cdot(\anchor_T- \anchor) > \prunethreshold$, where $\anchor_T$ is the maximum size on which we can train, then we have evidence that the learner is not relevant.

The problem is of course that we can compute neither $\lc(\anchor)$ nor $\lc'(\anchor)$, because it refers to the \emph{true} learning curve.
That is, \lc realizes Eq. (\ref{eq:risk}) for the hypotheses produced by a learner on different sample sizes.
So we can only try to \emph{estimate} the values of \lc and $\lc'$ at selected anchor points.

Given a sequence of anchor points $\anchor_1,..,\anchor_t$ for which we already observed performance values of a learner, we need to estimate $\lc(\anchor_t)$ and $\lc'(\anchor_t)$ in order to conduct the prune check for the performance at the full dataset size. 
A canonical estimate of $\lc(\anchor_t)$ is the empirical mean $\empmean_t$ of the observations at $\anchor_t$.
For the slope, we could rely on these estimates and use convexity to bound the slope with
\begin{equation}
\begin{aligned}
    \label{eq:slopeestimate:indirect}
    \lc'(\anchor_t) = \lc(\anchor_t) - \lc(\anchor_t-1) & \overset{conv}{\geq} \frac{\lc(\anchor_t) - \lc(\anchor_{t-1})}{\anchor_t - \anchor_{t-1}}\\
    &\approx \frac{\empmean_t - \empmean_{t-1}}{\anchor_t - \anchor_{t-1}}
\end{aligned}
\end{equation}

However, replacing $\lc'$ in the above decision rule by this bound is admissible \emph{only if} the estimates $\empmean_t$ and $\empmean_{t-1}$ are accurate (enough).
This calls for the usage of \emph{confidence intervals} to obtain a notion of probabilities (here, from a frequentist viewpoint, the probability to observe a mean close to the true one).
To compute such confidence bounds, we follow the common assumption that observations at each anchor follow a normal distribution \cite{domhan2015speeding,klein2017learning,kleinfastbo2017}.
Since the true standard deviation is not available for the estimate, we use the empirical one.

Collecting enough samples to shrink such confidence intervals to a minimum can be prohibitive, so we propose an alternative method, which estimates the slope in Eq. (\ref{eq:slopeestimate:indirect}) optimistically from the confidence bounds.
To this end, we connect the upper end of the interval at $\anchor_{t-1}$ with the lower end of the interval of $\anchor_t$.
This gives us the slope of the most \emph{optimistic} learning curve compatible with the observed performances up to a reasonable probability.
Formally, let $C_i$ be the confidence intervals at anchor $\anchor_i$.
Extrapolating from the \emph{last} anchor $\anchor_t$, the best possible performance of this particular learner at the final anchor $\anchor_T$ is
\begin{equation}
    \lc(\anchor_T) \geq \inf C_t - (\anchor_T-\anchor_t)\left(\frac{\sup C_{t-1} - \inf C_t}{\anchor_{t-1} - \anchor_t}\right),
\end{equation}
which can now be compared against the performance of the best evaluated learner so far \prunethreshold to decide on pruning.

\subsection{Calibrating the Width of Confidence Intervals}
\label{sec:lccv:convexitycompatibleobservationsets}
Given the above approach based on confidence intervals, the next question is on how many samples these intervals should be based on.
Clearly, at least two observations are needed to obtain a standard deviation and hence to create a meaningful interval.
However, apart from this restriction, the number of samples per anchor is a factor that is under our control.
We should determine it carefully as a too low number might imply inaccurate or large confidence intervals while a high number would maybe unnecessarily slow down the validation process.

The behaviour of LCCV can here be controlled via four parameters.
The first and the second parameters are simply the minimum and the maximum number of evaluations at each anchor point, respectively.
A minimum of other than 2 can make sense if we either want to allow greediness (1) or to be a bit more conservative about the reliability of small samples ($>2$).
To enable a dynamic approach between the two extremes, we consider a stability criterion \maxciwidth.
As soon as the confidence interval width drops under \maxciwidth, we consider the observation set to be stable.
Since error rates reside in the unit interval in which we consider results basically identical if they differ on an order of less than $10^{-4}$, this value could be a choice for \maxciwidth.
While this is indeed an arguably good choice at the last anchor, the ``inner'' anchors do not require such a high degree of certainty.
The goal is not to approximate the learning curve at all anchor points with high precision but only to get a rough grasp on its shape.
To achieve this objective, a loose confidence interval of size even 0.1 might be perfectly fine.
Otherwise, the validation routine might spend a lot of time on ultimately not so relevant anchors.
Hence, we distinguish between the standard \maxciwidth and a parameter $\maxciwidth_{max}$ applied only for the last anchor.
Appendix~\ref{sec:sensitivityanalysis} in the supplementary material provides a sensitivity analysis over these hyperparameters.

Since the stability criterion \maxciwidth could be rather loose, it makes sense to adopt an additional sanity check that assures that the observation set is \emph{convexity-compatible}.
We call a set of observations convexity-compatible if there exists a convex function that takes at each anchor point a value that is in the confidence interval of the respective anchor.
In other words, we do not consider a single empirical learning curve but \emph{all} (convex) learning curves that are compatible with the current confidence bounds at the anchors.

There is an easy-to-check necessary and sufficient condition of the convexity-compatible property.
To this end, one tries to construct a curve backward starting from the last two intervals and to decrease the slope as little as possible from each anchor point to the next one.
Formally, let $C_1,..,C_t$ be the confidence intervals at the $t$ ascendingly ordered anchor points.
We set $u_t = \sup C_t$ as the \emph{worst} possible learning curve value at the \emph{last} anchor point and then consider a minimum possible negative slope $\sigma_i$ at each anchor $i$.
Starting with $\sigma_t = 0$, we can inductively define
\begin{equation}
    \begin{aligned}
        u_{i-1} & = \min\{x\in C_{i-1}~|~x \geq u_i - (\anchor_i - \anchor_{i-1})\sigma_i\}~~~\text{ and }\\
        \sigma_{i-1} &= \frac{u_i - u_{i-1}}{\anchor_i - \anchor_{i-1}}
    \end{aligned}
\end{equation}
Starting off with the worst value in the last confidence interval is important to impose the least possible slope in the beginning, since the slope must keep (negatively) increasing.

If the set from which $u_{i-1}$ is chosen is non-empty for all anchor points, we have constructed a convex curve inside the confidence bands  (necessary condition), and it is also easy to see that there cannot be a convex curve if we fail to produce one in this way (sufficient condition).
In other words, one can simply check for convexity compatibility by checking whether the sequence of slopes of the lines that connect the upper bound of $C_i$ with the lower bound of $C_{i+1}$ is \emph{increasing}.

If an observation set is not convexity compatible, we need to ``repair'' it in order to be able to prune in a legitimate way.
Our repair technique is to simply step back one anchor.
Stepping back from anchor $s_t$ to $s_{t-1}$, the algorithm will typically only re-sample once at anchor $s_{t-1}$ since the confidence bound is already small enough.
Two cases are possible now.
Either the convexity-compatibility now is already lost at anchor $s_{t-1}$, in which case the same procedure is applied recursively stepping back to anchor $s_{t-2}$.
Otherwise, it will return to anchor $s_t$ and also re-sample a point at that anchor.
In many cases, the convexity compatibility will have recovered at this time.
If not, the procedure is repeated until the problem has been resolved or the maximum number of evaluations for the anchors has been reached.

\subsection{Skipping Intermediate Evaluations}
\label{sec:lccv:skippingevaluations}
Unless we insist on evaluating at all anchor points for obtaining insight about the learning process, we should evaluate a learner on the full data once it becomes evident that it is definitely competitive.
This is obviously the case if we have a \emph{stable} (small confidence interval) anchor point score that is at least as good as the best observed performance \prunethreshold; in particular, it is automatically true for the first candidate.
Of course, this only applies to non-incremental learners; for incremental learners, we just keep training to the end or stop at some point if the condition of Sec. \ref{sec:lccv:cancellation} applies.

Note that, in contrast to abortion, we cannot lose a relevant candidate by jumping to the full evaluation.
We might waste some computational time, but, unless a timeout applies, we cannot lose relevant candidates.
Hence, we do not need strict guarantees for this decision.

With this observation in mind, it indeed seems reasonable to \emph{estimate} the performance of the learner on the full data and to jump to the full dataset size if this estimate is at least as good as reference value \prunethreshold.
A dozen of function classes have been proposed to model the behaviour of learning curves (see, e.g.,~\cite{domhan2015speeding}), but one of the most established ones is the inverse power law (IPL) \cite{nonlinearregression}.
The inverse power law allows us to model the learning curve as a function
$\hat \lc(\anchor) = a - b\anchor^{-c}$,
where $a,b,c \in \mathbb{R}_{+}$.
Given observations for at least three anchor points, we can fit a non-linear regression model using, for example, the Levenberg-Marquardt algorithm \cite{nonlinearregression}.
After the sampling at an anchor has finished, we can fit the parameters of the above model and check whether $\hat \lc(\anchor_T) \leq \prunethreshold$.
In that case, we can immediately skip to the final anchor. 

\subsection{The LCCV Algorithm}
\label{sec:lccv:algorithm}
The LCCV algorithm is sketched in Alg. 1 and puts all of the above steps together.
The algorithm iterates over anchors in a fixed schedule \schedule (in ascending order).
At anchor point \anchor (lines \ref{algline:stage:start}-\ref{algline:stage:end}), the learner is validated by drawing folds of training size \anchor, training the learner on them, and computing its predictive performance on data not in those folds.
These validations are repeated until a stopping criterion is met (cf. Sec.  \ref{sec:lccv:convexitycompatibleobservationsets}).
Then it is checked whether the observations are currently compatible with a convex learning curve model.
If this is not the case, the algorithm steps back one anchor and gathers more observations in order to get a better picture and ``repair'' the currently non-convex view (l. \ref{algline:repair}); details are described in Sec. \ref{sec:lccv:convexitycompatibleobservationsets}.
Otherwise, a decision is taken with respect to the three possible interpretations of the current learning curve.
First, if it can be inferred that the performance at $\anchor_T$ will not be competitive with the best observed performance \prunethreshold, LCCV returns \void (l. \ref{algline:prune}, cf. Sec. \ref{sec:lccv:cancellation}).
Second, if extrapolation of the learning curve gives rise to the belief that the learner is competitive, then LCCV directly jumps to validation on the full dataset of the candidate (l. \ref{algline:skip}, cf. Sec. \ref{sec:lccv:skippingevaluations}).
In any other case, we just keep evaluating at the next anchor (l. \ref{algline:standardcase}).
In the end, the estimate for $\anchor_T$ is returned together with the confidence intervals for all anchor points.

\input{lccv-algo}

We observe that the runtime of LCCV is at most twice as much as the one of kCV in the worst case, if we assume a computational complexity that is at least linear in the number of data points (proof in Appendix~\ref{app:proof} of the supplement):
\begin{theorem}
Let \dataset be a dataset, $T = \log_2|\dataset|$ and $\rho \in \mathbb{N}, \rho \leq T$.
For anchor points $S = (2^\rho,2^{\rho+1}..,2^{\floor{T}}, 2^T)$ and with a maximum number of $k$ samples per anchor point, the runtime of LCCV is at most twice as high as the one of kCV, where $2^d$ is the training size applied by kCV.
\end{theorem}

A runtime bound of twice the runtime of kCV seems not very exciting.
However, this is a \emph{worst case} bound, and empirical results will show that most of the time LCCV is even faster than kCV due to its pruning mechanism.
Besides, the quality of insights obtained from LCCV substantially augment what is learned from kCV, because it allows us to make recommendations to the user of whether to acquire more instances or to find better features; we discuss this in more detail in Sec. \ref{sec:evaluation:recommendations}.

\section{Evaluation}
This evaluation addresses two main questions:
\begin{enumerate}
    \item How does the runtime performance of a model selection process employing LCCV compare to the usage of 5CV and 10CV, and does the usage of LCCV produce competitive results?
    These questions are considered in Sec. \ref{sec:evaluation:randomsearch}.
    
    \item Can we use the collected empirical learning curves to make correct recommendations on the question of whether more data will improve the overall result?
This is subject of Sec. \ref{sec:evaluation:recommendations}.

\end{enumerate}

To enable full reproducibility, the implementations of all experiments conducted here alongside the code to create the result figures and tables, detailed log files of the experiment runs, and result archives are available for the public\footnote{\url{github.com/fmohr/lccv/tree/master/publications/2022TPAMI}}.

\subsection{LCCV for Model Selection}
\label{sec:evaluation:randomsearch}
In this evaluation, we compare the effect of using LCCV instead of 5CV or 10CV as validation method inside a simple \automl tool based on random search.

\subsubsection{Benchmark Setup}
Our evaluation measures the runtime of a random search \automl tool that evaluates $200$ classification pipelines from the scikit-learn library \cite{pedregosa2011scikit} using different validation techniques to assess candidate performance.
The pipelines consist of a classification algorithm, which may be preceded by up to two pre-processing algorithms.
The behaviour of the random search is simulated by creating a sequence of $200$ random pipelines (including the hyperparameter values for the contained algorithms); the considered algorithms and the exact procedure are explained in Appendix~\ref{sec:appendix:pipelines} of the supplement and the resulting pipelines are published in the resources.
The model selection technique iterates over this sequence, in each iteration it calls the validation method for the candidate and updates the currently best candidate based on the result.
For LCCV, the currently best observation is passed as parameter \prunethreshold, and a candidate is discarded if \void is returned.

We compare 5CV and 10CV with two different configurations of LCCV.
When comparing with 5CV, we suppose that the goal is to find the pipeline with the best performance estimate when using 80\% of the data for training.
For LCCV, we set $s_T := 0.8|\dataset|$ for a dataset \dataset and dub this 80LCCV.
When comparing with 10CV, we suppose that the goal is to optimize for training on 90\% of the data, set $s_T := 0.9|\dataset|$ and dub this 90LCCV.
That is, we conduct two pairwise comparisons, once comparing 5CV against 80LCCV and once 10CV against 90LCCV.

The \emph{runtime} of a validation method on a given dataset is the average total runtime of the random search using this validation method.
Of course, the concrete runtime can depend on the concrete set of candidates but also, in the case of LCCV, their order.
Hence, over 10 seeds, we generate different classifier \emph{portfolios} to be validated by the techniques, measure their overall runtime (of course using the same classifier portfolio and identical order for both techniques), and form the average over them.

The \emph{performance} of the chosen learner of each run is computed by an exhaustive MCCV.
To this end, we form 100 bi-partitions of split size 90\%/10\% and use the 90\% for training and 10\% for testing in all of the 100 repetitions (the split is 80\%/20\% when comparing 5CV against 80LCCV).
The average error rate observed over these 100 partitions is used as the validation score.
Note that we do not need anything like ``test'' data in this evaluation, because we are only interested in whether the LCCV can reproduce the same model selection as kCV.
More precisely, we do not try to estimate the true learning curve value at the target train size but the value one would obtain in expectation when using only the \emph{available} data.
While this value tends to be an optimistically biased estimate of the true value, the true value itself is just not of interest for the validation mechanism.
This is a problem that falls into the responsibility of an anti-over-fitting mechanism, which is not part of the validation mechanism and hence not part of the experiments.

As common in \automl, we configured a timeout per validation.
In these experiments, we set the timeout per validation to 5 minutes.
This timeout refers to the \emph{full} validation of a learner and not, for instance, to the validation of a single fold.
Put differently, 5 minutes after the random search invoked the validation method, it interrupts validation and receives a partial result.
For the case of kCV this result is the mean performance among the evaluated folds or \texttt{nan} if not even the first fold could be evaluated.
For LCCV, the partial result is the average results of the biggest evaluated anchor point if any and \texttt{nan} otherwise.

To obtain insights over different types of datasets, we ran the above experiments for all of the 67 datasets of the \automl benchmark suite \cite{gijsbers2019open}.
These datasets offer a broad spectrum of numbers of instances and attributes.
All the datasets are published on the \url{openml.org} platform \cite{feurer2021openml,OpenML2013}.
As a means of general pre-processing, missing values are replaced by the median for numerical and the mode for categorical attributes.
After that, all categorical attributes are binarized. 

\subsubsection{Parametrization of LCCV}
LCCV was configured as follows.
Following the behaviour of kCV, the upper sample limit per anchor is 5 for 80LCCV and 10 for 90LCCV.
The minimum number of samples was set to 3.
The minimum exponent \minexp was set to 6 to train over at least 64 instances.
The maximum width parameter \maxciwidth for the confidence intervals was set according to the anchor.
For the last anchor of the maximum exponent it was configured with $\maxciwidth_{max} = 0.001$.
For intermediate anchors it was configured with $\maxciwidth = 0.1$.
We have conducted a sensitivity analysis for some of the parameters, which can be found in Appendix~\ref{sec:sensitivityanalysis} of the supplementary material.

\subsubsection{Technical Specification}
The computations were executed in a compute center with Linux machines, each of them equipped with 2.6Ghz Intel Xeon E5-2670 processors and 20GB memory.
In spite of the technical possibilities, we did \emph{not} parallelize evaluations in the interest of minimizing potential confounding factors.
That is, all the experiments were configured to run with a single CPU core.

\subsubsection{Results}
\begin{figure}[t]
    \centering
    \includegraphics[width=\columnwidth]{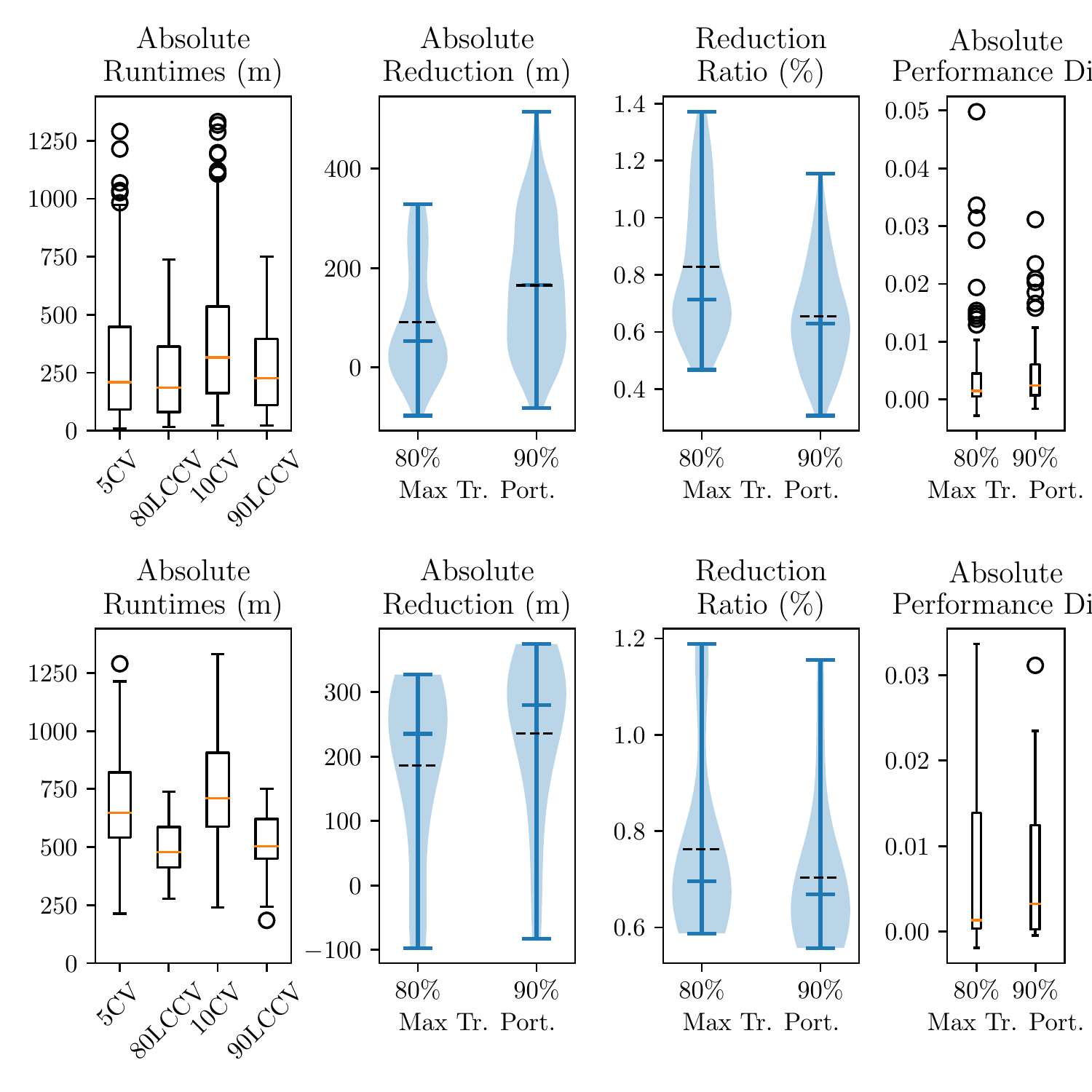}
    \caption{Comparison between LCCV and kCV as validators in a random search.
    Top panel: results over all the 67 datasets.
    Bottom panel: results over the 19 datasets with runtimes of at least 6 hours.}
    \label{fig:results:randomsearch:boxplots}
\end{figure}

The results are summarized in Fig. \ref{fig:results:randomsearch:boxplots}.
We first consider the top row.
The boxplots capture, in this order, the average runtimes of LCCV and kCV on each of the 67 datasets (leftmost figure), the absolute and relative \emph{reductions} of runtimes in minutes when using LCCV compared to kCV (central plots), and the deviations in the error rate of the eventually chosen model when using LCCV compared to kCV (rightmost figure).
While the absolute reduction (middle left) is in terms of saved minutes (more is better from the LCCV viewpoint), the relative runtime reduction refers to the \emph{ratio} between the runtime of LCCV and kCV (less is better from the LCCV viewpoint).
More detailed insights, e.g., visualizations per dataset, can be found in Appendix~\ref{sec:appendix:results:randomsearch} of the supplementary material.

The first observation is that LCCV comes with a useful average runtime reduction over kCV.
Looking at the absolute reductions (top panel, left centre plot), we see that the mean reduction (dashed line) is around 91 minutes over 5CV and 165 minutes over 10CV.
More precisely, this is an average absolute reduction from 5.5h for 5CV to 4h for 80LCCV and from roughly 7h for 10CV to 4.5h for 90LCCV.
In relative numbers, the runtime of 80LCCV is only 83\% the runtime of 5CV and the one of 90LCCV is only 65\% of the runtime of 10CV on average.
The median reduction ratios are even better than that.

Of course, the time savings are less relevant on small datasets but all the more impacting on the bigger ones.
The bottom panel of Fig. \ref{fig:results:randomsearch:boxplots} shows the same results but only for the 19 out of the 67 datasets on which the runtime of 80LCCV is above 6h.
If limiting the analysis to those datasets, the absolute average runtime reduction is from 11h for 5CV (a reduction to 76\%) to 8h for 80LCCV and from 12h for 10CV to 8h for 90LCCV (a reduction to 70\%).

LCCV does not always improve runtime.
There is a handful of cases in the comparison between 5CV and 80LCCV in which 80LCCV is slightly slower (time difference of under 10 minutes) but also two cases of runtime differences of an hour or more.
In any case, the runtimes are \emph{far} away from the worst-case factor of 2 derived in Sec. \ref{sec:lccv:algorithm}.
The increase occurs in fact due to a higher number of evaluations that, in total, exceeds the time that would have been required for evaluating using all data right away.
In particular, we found LCCV to be inefficient on datasets with few instances and many attributes like {\sc Amazon}.
While intuition already suggests that building learning curves for model selection is not beneficial in those cases, the learning curves are all the more important here for dataset management and in order to estimate potential gains with more instances as discussed in Sec. \ref{sec:evaluation:recommendations}.
This can be considered the price for the additional insights produced with LCCV as will be discussed in Sec. \ref{sec:evaluation:recommendations}.
In the comparison between 10CV and 90LCCV this effect does almost not occur.

Increased runtime can also have negative effects on the performance.
In fact, the performance of LCCV can even get \emph{worse} in such cases.
For example, in the latter two cases of substantially increased runtime of 80LCCV, we also observe substantially worse results compared to 5CV.
The reason for this behaviour is that, in those cases, 80LCCV eventually neither prunes nor skips and tries to evaluate all anchor points (high runtime) and then often runs into the timeout and cuts the candidate that would be competitive (bad performance).
This is clearly an undesirable situation that should be avoided.
Specifically in the timeout scenarios, one can hope to resolve the problem with a different parametrization of LCCV, but it is clearly desirable to achieve higher robustness that does not depend on parameters in future work.

The final observation is dedicated to the last plot and the performance loss we need to pay by using LCCV instead of kCV.
In 90\% of the cases, the performance of LCCV is in a range of 0.015 of the one of kCV, and in 83\% of the cases, the gap is even under 0.01.
These numbers are encouraging and, in fact, the performance differences are negligible in most of the cases.
Nevertheless, from a strict viewpoint, LCCV falls a bit short with respect to the constraints defined in Eq. (\ref{eq:constraintonsmartvalidation}) of Sec. \ref{sec:problem}, because it occasionally underperforms the kCV.
There is no general explanation that covers all cases.
We have above discussed the problem of timeouts on optimal candidates, which explains some of the cases.
Another explanation can be illegal pruning due to situations in which the convexity assumption is violated.

To summarize, in our interpretation, LCCV is in spite of some exceptions largely superior to kCV with respect to runtimes.
LCCV is particularly strong on the ``hard'' datasets with high training runtimes.
While LCCV is also competitive in terms of performance most of the time, it occasionally underperforms kCV, which gives rise to improvements of the LCCV technique in future work.

\subsection{LCCV for Data Acquisition Recommendations}
\label{sec:evaluation:recommendations}
We analyze whether the promise of LCCV to give more insights than kCV is justified.
We use the produced empirical learning curves to derive an IPL model with which we answer the question of whether more instances will lead to \emph{sufficiently} better results or not.
Of course, in practice, we will be interested not in any marginal improvement at any cost but will require a reasonable ratio between additional instances and improvement.

\subsubsection{Research Question}
We want to assess how suitable the empirical learning curves collected with LCCV are for predicting whether or not more data would be useful for the user.
While more instances typically imply better results, additional instances also come at additional costs; this question is hence not trivial.
In the most general case, we can define a \emph{utility} function over the sample size and try to find the optimal sample size \cite{weiss08maximizingclassifierutility,last2009improving}.
Since the return on investment decreases for increasing sample sizes, the utility assumes its global optimum at the \emph{economic stopping point} and strictly decreases thereafter.

Since the concrete utility function depends on application contexts,  we adopt a simplifying approach in this paper to address the above question.
Instead of introducing a concrete utility function, which is everything but trivial \cite{weiss08maximizingclassifierutility}, we reduce the problem to a binary decision situation:
The user has the option to double the number of available instances and can choose to do so or not.
Since there are only two possible situations, we only need to express a condition under which accepting the additional instances is advantageous over staying with the current set.
To this end, a threshold \utilitythreshold expresses how much the performance (here error rate) must improve in order to have a utility gain by accepting the additional data.
In other words, the utility function is, ordinally and only for the two possible states, characterized by \utilitythreshold.

Importantly, the question about improvement does typically not refer to a specific learner but to the improvement that \emph{any} learner of a considered portfolio could achieve over the \emph{currently best} one if more data was available.
That is, we are not necessarily interested in whether a learner can improve by at least \utilitythreshold on its own result but whether the result of the \emph{best} learner can be improved by at least \utilitythreshold by the \emph{portfolio}.
Note that this is not the same question, because it could be that the currently best learner has a stalling learning curve but that a runner-up learner has a still dropping learning curve and would take the lead given that more instances were available.

We answer the following two research questions:
Modelling learning curves with the inverse power law (IPL, cf. Sec. \ref{sec:lccv:skippingevaluations}) fitted from the empirical curves produced by LCCV,
\begin{enumerate}
    \item how well can we predict the \emph{performance improvement} (regression problem) of both each learner and a whole portfolio when the available data is doubled?
    
    \item how well can we predict, for different values of \utilitythreshold, the utility-maximizing decision (binary classification problem)?
\end{enumerate}

\subsubsection{Realization}
To assess the performance of a recommender for these questions on a dataset \dataset, we hold a substantial portion of \dataset apart that will then serve as ``new'' data.
Of course, new data here does not mean untouched validation data but additional training instances that are pretended to be not available before.
First, we sample two subsets of the data, \datasetfull and \datasethalf, of size $|\datasetfull| =  10/9 \cdot 2^{\floor{\log_2(0.9|\dataset|)}}$, and $|\datasethalf| = 10/9\cdot 2^{\floor{\log_2(0.9|\dataset|)} - 1}$, respectively, where
\datasethalf is sampled as a subset from \datasetfull.
These perhaps weird-looking sizes are chosen so that using 90\% of the data for training will correspond to a power of 2; here the highest power(s) possible for the train set of the dataset.
Then, using only the data of \datasethalf, we validate all classifiers in the portfolio with LCCV without cancellation and skipping (cf. Sec. \ref{sec:lccv:cancellation}, \ref{sec:lccv:skippingevaluations}), so that we have full evaluations at all anchor points; the runtime is still bound by a factor of two (cf. Sec. \ref{sec:lccv:algorithm}).

To make and qualify a prediction of whether doubling the number of instances will yield an improvement by at least \utilitythreshold, we proceed as follows.
We denote the performance of algorithm $i$ trained on 90\% and validated against 10\% of the data in \datasethalf and \datasetfull as $P^i_{\datasethalf}$ and $P^i_{\datasetfull}$, respectively. 
The performance of the \emph{best} performing algorithm is denoted as $P^*_{\datasethalf}$ for \datasethalf and $P^*_{\datasetfull}$ for \datasetfull.
Pretending that \datasethalf is the actually available data, the quantities $P^i_{\datasethalf}$ and hence $P^*_{\datasethalf}$ are known after having applied LCCV to the portfolio, but $P^i_{\datasetfull}$ and $P^*_{\datasetfull}$ are \emph{unknown} and must be estimated.
The correct answer is ``yes'' if $P^*_{\datasethalf} - P^*_{\datasetfull} \geq \utilitythreshold$ and ``no'' otherwise. 
Given the learning curves of a portfolio of classifiers validated with LCCV, an IPL model (cf. Sec. \ref{sec:lccv:skippingevaluations}) is built for each of them using non-linear regression; this model is called the \emph{IPL-regressor}.
We denote the \emph{predicted} performance of this model for algorithm $i$ on \datasetfull as $\hat P^i_{\datasetfull}$.
Then, we compute $\hat P^*_{\datasetfull} := \min_i \hat P^i_{\datasetfull}$ and return ``yes'' if this value is at most $P^*_{\datasethalf} - \utilitythreshold$ and ``no'' otherwise.
The implementation of this last decision rule is called the \emph{IPL-classifier} for threshold \utilitythreshold.

In this evaluation, we consider a simple portfolio of 17 classifiers.
These are the same classifiers from scikit-learn library \cite{pedregosa2011scikit} as considered for the pipelines in Sec. \ref{sec:evaluation:randomsearch}:
BernoulliNB, GaussianNB, Decision Tree, Extra Trees (ensemble), Random Forest,
GradientBoosting, kNN, SVC (with four different kernels), MLP, MultinomialNB, PassiveAggressive, LDA, QDA, SGD.
We refer to Appendix~\ref{sec:appendix:algorithms} of the supplement for details.

\begin{figure*}[t]
    \centering
    \includegraphics[width=\textwidth]{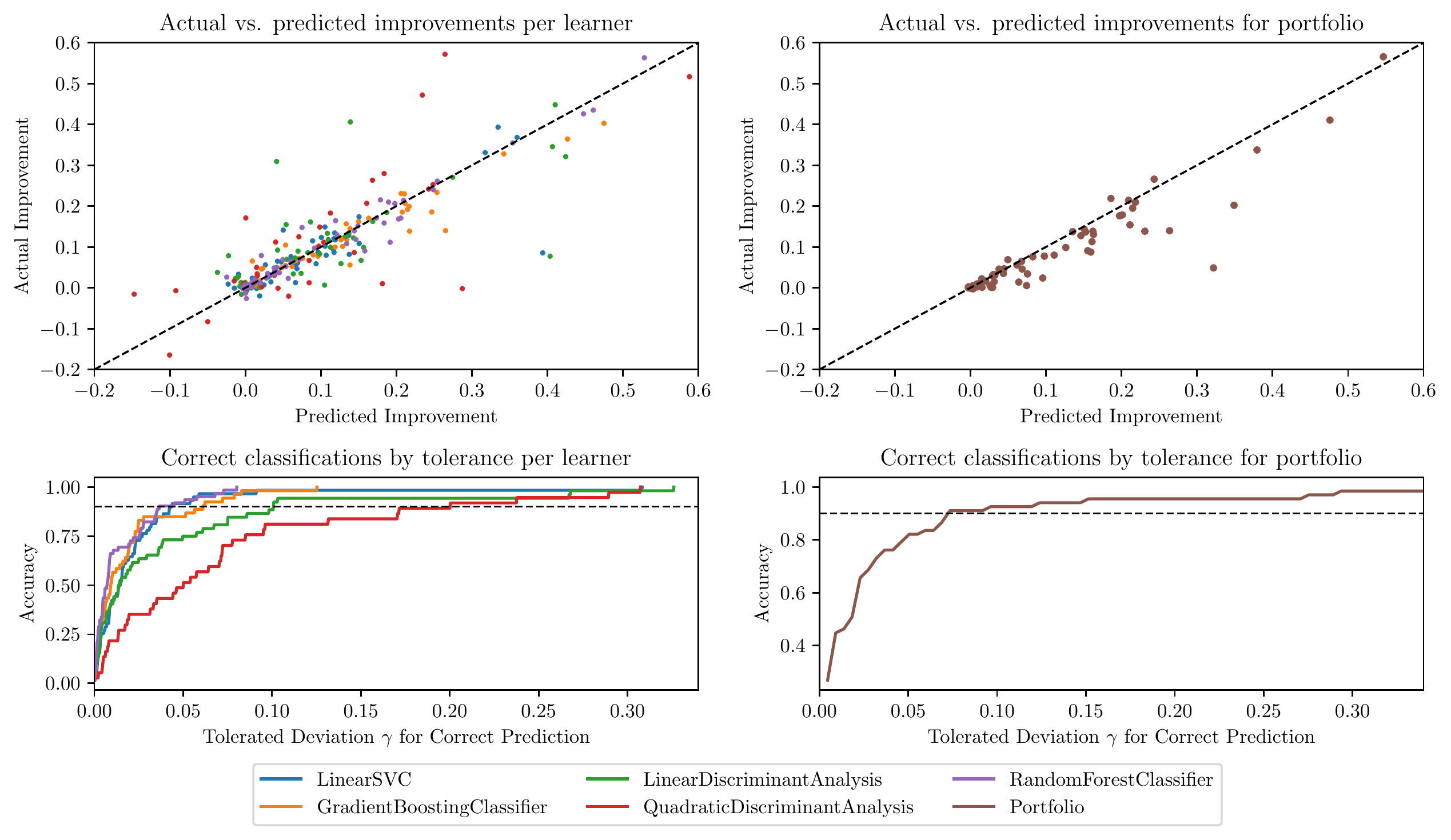}
    \caption{Correlation between the extrapolated learning curve and the actual prediction. \label{fig:extrapolation-correlation}}
\end{figure*}

\subsubsection{Results}

\begin{figure*}[t]
    \centering
    \includegraphics[width=.95\textwidth]{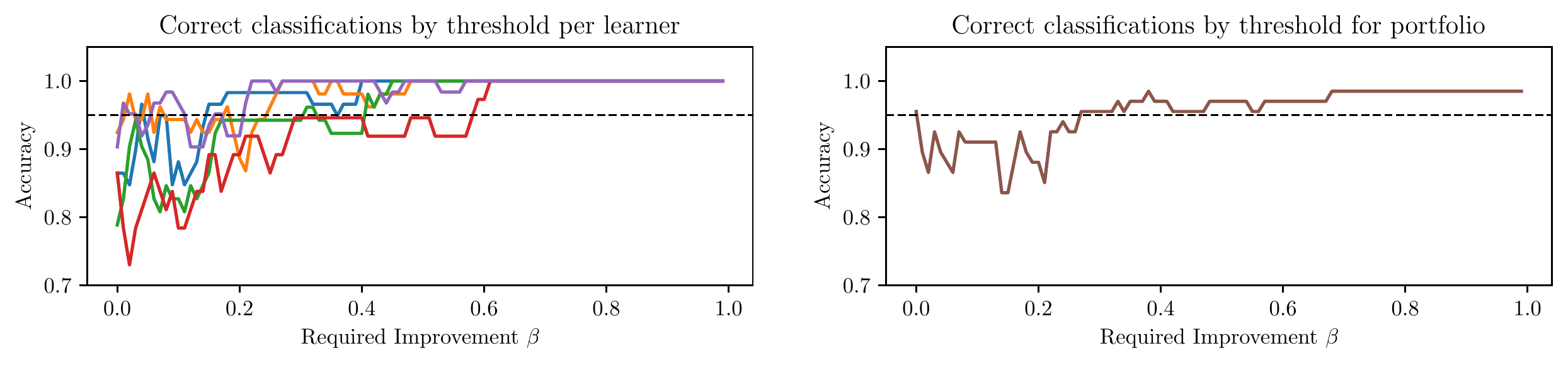}
    \caption{Accuracy of the inverse power law model extrapolating the learning curve.
    \label{fig:extrapolation-accuracy}}
\end{figure*}

The results are presented in Fig.~\ref{fig:extrapolation-correlation} and Fig.~\ref{fig:extrapolation-accuracy}. 
In both figures, results in the left column refer to (actual and predicted) improvements for individual classifiers over themselves, and the results in the right column refer to improvements over the \emph{best} classifier in the portfolio.
For the left column plots, we have included only the insights of 5 out of the 17 classifiers for readability, which are Linear SVC (blue), Gradient Boosting (orange), LDA (green), QDA (red), Random Forest (purple).

To assess the first research question, we consider Fig.~\ref{fig:extrapolation-correlation}.
We first focus on the top row of the figure, which shows the correlation between the predicted improvement and the actual improvement. 
Each data point represents the performance of an algorithm on a dataset; all points of the same colour belong to the same algorithm.
The x-value of a point is the \emph{predicted improvement} for an algorithm $i$ when increasing the data from \datasethalf to some \datasetfull with twice as much instances, i.e. $P^i_{\datasethalf} - \hat P^i_{\datasetfull}$.
The y-axis shows the \emph{actual improvement}, i.e. $P^i_{\datasethalf} - P^i_{\datasetfull}$.
Hence, points close to the diagonal indicate reliable predictions.
In the right plot, we show how this generalizes towards the portfolio setting.
That is, the x-axis represents the \emph{predicted} improvement of the \emph{portfolio}, i.e. $P^*_{\datasethalf} - \hat P^*_{\datasetfull}$, and the y-axis representing the \emph{true} improvement of the portfolio, i.e. $P^*_{\datasethalf} -  P^*_{\datasetfull}$.
Here, the portfolio consists of all 17 classifiers.

The correlation plots show that for most classifiers the predicted improvement is correlated to the actual improvement. 
The points of the various classifiers are evenly spread around the diagonal, indicating that the individual learning curves of the various algorithms are predictable. 
This was also mentioned by various other authors, e.g.,~\cite{domhan2015speeding,klein2017learning}. 
The predictions per portfolio are often optimistic. 

The bottom row of Fig.~\ref{fig:extrapolation-correlation} shows a more aggregated view on the same data.
For different values of the tolerance parameter \toleranceparam (x-axis), it shows the \emph{portion} of points whose discrepancy between the predicted and the true value is at most \toleranceparam; one could say that it shows the ``accuracy'' of the IPL-\emph{regressor} with respect to tolerance \toleranceparam.
Formally, the y-axis show the percentage of cases where $|P^i_{\datasethalf} - \hat P^i_{\datasetfull}| \leq \toleranceparam$ for the left plot and $|P^*_{\datasethalf} - \hat P^*_{\datasetfull}| \leq  \toleranceparam$ for the right plot. 
Relating this graph to the top row, it shows the relative frequency of points whose horizontal distance to the dashed line in the upper plot is at most \toleranceparam.
Note that these curves are by definition strictly increasing.
The dashed horizontal line is a visual aid to indicate the 90\% threshold.

The interpretation of the here interesting right plot is as follows.
If we accept a mistake of up to 0.025 for the prediction of \emph{how much} we can improve over the best learner at \datasethalf when doubling the data, we will get a ``correct'' answer in roughly 50\% of the cases.
If we accept a mistake of 0.07 or more instead, the answer will be correct in over 90\% of the cases.

With respect to the research question under consideration, namely, whether we can well predict the improvement achieved when doubling the data, the results are, depending on the expectations, a bit mediocre.
On the positive side, the model is precise up to 0.01 on over 40\% of the analyzed cases.
On the downside, being precise up to 0.025 (which may be an acceptable range) in only 50\% of the cases might seem an unacceptable gamble.
It is worth mentioning that better results might be achievable under other learning curve models than the IPL, e.g. using exponential laws \cite{gu2001modelingclassificationperformance}.

However, for the decision situation at hand, the above precision is unnecessarily strict.
In fact, being able to accurately predict the learning curve value for data acquisition questions is only important if (i) we can control the exact number of instances to be gathered and (ii) indeed have a very detailed concept of utility, which is often not the case in practice.
In our decision situation, we only need to know whether, based on the recommendation, we take the correct \emph{decision} in accepting the additional batch of data or not.
In other words, if the true improvement is much more than \utilitythreshold and we predict just a value slightly above \utilitythreshold, then we might make a rather big mistake from the regression point of view but \emph{no} mistake from the classification point of view.
The same holds for values below \utilitythreshold.
So one might argue that the prediction error is not so severe as long as we act in the correct way.

Inversely, great regression performance does not even necessarily imply substantially better decisions.
If we make a mistake in the performance prediction, then the crucial question is whether this mistake will lead to a decision that is different from the decision one should take if the performance was \emph{known}.
But if the true performance improvement is very close to \utilitythreshold, then even a very small mistake in the performance prediction can lead to a wrong decision.
This fact also explains the ``oscillation'' in Fig.~\ref{fig:extrapolation-accuracy}, which we discuss below in more depth; a similar effect was described in \cite{mohr2021predicting} when predicting whether the execution of a learning algorithm will time out.
To summarize, both performances (classification and regression) are of interest and even though great regression results would be desirable, neither are they necessary nor sufficient for good classification performance.

Motivated by these observations, we can address the second research question with the plots in Fig.~\ref{fig:extrapolation-accuracy}.
This figure shows the accuracy of the IPL-\emph{classifier} (y-axis) when the task is to predict whether the double amount of data will improve the performance at least by a certain percentage point \utilitythreshold (x-axis).
On the left, this is the percentage of cases where $(P^i_{\datasethalf} - \hat P^i_{\datasetfull} < \utilitythreshold)\equiv(P^i_{\datasethalf} - P^i_{\datasetfull} < \utilitythreshold)$ holds.
In other words, the fraction of cases where the recommended action is indeed the one that maximizes the utility.
In the right plot, this is done for the portfolio, i.e., the percentage of cases where $(P^*_{\datasethalf} - \hat P^*_{\datasetfull} < \utilitythreshold) \equiv (P^*_{\datasethalf} -  P^*_{\datasetfull} < \utilitythreshold)$ holds. 
The dashed line is a visual aid for an accuracy score of 95\%.

Maybe the first observation here is that, in contrast to the plots in Fig. \ref{fig:extrapolation-correlation}, the curves in this figure are not monotone.
The reason for this is precisely that the response is likely to be correct if the required improvement \utilitythreshold is very low (close to 0) or very high (close to 1).
In the first case, always predicting that the improvement will be achieved will most of the time be correct while predicting that it will not be achieved for high values of \utilitythreshold is likewise a safe prediction.
Hence, the difficult cases are those in which the true improvement is, incidentally, close to the required threshold \utilitythreshold.
In fact, in those cases, even a very small prediction error can cause a wrong binary prediction.
For this reason, the accuracy drops a bit in the range of $\beta \in [0.05, 0.2]$.

However, these performance drops are still at a quite high level and much better than what the regression task suggested.
Looking at the right plot for the portfolios, we can see that for \emph{any} value \utilitythreshold of a required improvement over the \emph{best} learner, we predict with over 80\% accuracy correctly whether or not such an improvement will be realized when doubling the data.
For most thresholds, this score is even above 90\%.
We hence conclude that the IPL-classifier has a good probability to predict whether or not a required improvement \utilitythreshold will be realized \emph{regardless} the concrete value of \utilitythreshold and the true improvement.

To summarize, the empirical learning curves produced as a \emph{side product} of LCCV, give great insights into the learning behaviour of the considered algorithms and can be an important resource for decision making in data acquisition.
Even though the potential improvements of a portfolio cannot always be predicted with high precision, the predictions are mostly still sufficiently exact to recommend correctly about whether or not acquiring a fixed number of additional data points if the required improvement to obtain a positive net utility is defined.
While this type of analysis can at least partially also be realized by a-posteriori constructing learning curves of the best candidates in the portfolio according to a kCV, such a technique is not complete and could miss a learner that will turn the lead on more data while LCCV takes all the learners in the portfolio into account.

\section{Conclusion}
In this paper, we presented LCCV, a technique to validate learning algorithms based on learning curves.
In contrast to other evaluation methods that leverage learning curves, LCCV is designed to not prune candidates that can potentially still outperform the current best candidate.
Based on a convexity assumption, which turns out to hold almost always in practice, candidates are pruned once they are unlikely to improve upon the current best candidate. 
This makes LCCV potentially slower, but more reliable than other learning-curve-based approaches, whereas it is faster and mostly equally reliable as vanilla cross-validation methods. 

We have empirically shown that LCCV outperforms both 5CV and 10CV in many cases in terms of the runtime of a random-based model selection algorithm that employs these methods for validation.
Reductions are on the order of up to between 20\% and 50\%, which corresponds to 90 minutes reduction on average (160 minutes for 10CV) and up to over 5 hours in some cases.
We emphasize that LCCV is a contribution that is complementary to other efforts in \automl and can be used with many of the tools in this field.

Moreover, the learning curves can be used to give additional information to the data owner. 
When applying machine learning methods, data owners are often confronted with the choice of whether they should invest in getting more data, more attributes, or more model evaluations. 
LCCV has the ability to support the decision-maker.
We showed that it gives good confidence recommendations ($>80 \% $ accuracy) of whether the acquisition of new data will enable a given desired reduction of the error rate.

Future work should be concerned with several refinements of the technique.
First, it would be highly desirable to detect non-convexities to disable pruning in those cases.
Second, it would be interesting to adopt a more dynamic and curve-dependent manner of choosing anchor points instead of fixing these as constants before.
Finally, it would be interesting to integrate LCCV in various AutoML toolboxes, such as ML-Plan~\cite{mohr2018ml}, Auto-sklearn~\cite{feurer2015efficient}, and Auto-Weka~\cite{autoweka}.
Indeed, enabling these toolboxes to speed up the individual evaluations, has the potential to further push the state-of-the-art of machine learning research.

\footnotesize
\bibliographystyle{abbrvnat}
\bibliography{bibliography}

\begin{IEEEbiography}[{\vspace{-3em}\includegraphics[width=1in,keepaspectratio]{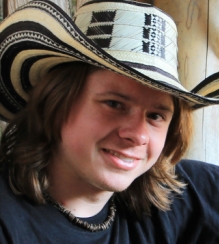}}]{Felix Mohr}
is a professor in the Faculty of Engineering at Universidad de la Sabana in Colombia.
His research focus lies in the areas of Stochastic Tree Search as well as Automated Software Configuration with a particular specialization on Automated Machine Learning.
He received his PhD in 2016 from Paderborn University in Germany.
\end{IEEEbiography}

\begin{IEEEbiography}[{\vspace{-3em}\includegraphics[width=1in,keepaspectratio]{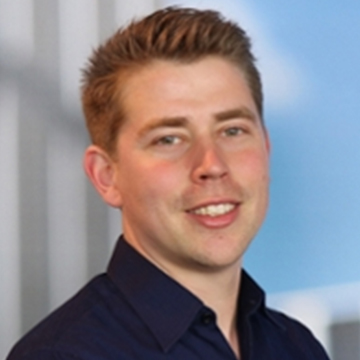}}]{Jan N. van Rijn} 
is assistant professor at Leiden Institute of Advanced Computer Science (LIACS), Leiden University (the Netherlands). During his PhD, he (co-)founded and developed OpenML, an open science platform for machine learning. After obtaining his PhD, he worked as a postdoctoral researcher in the Machine Learning lab at University of Freiburg (Germany), and Columbia University in the City of New York (USA).
His research interests include artificial intelligence, AutoML and metalearning.

\end{IEEEbiography}

\clearpage
\setcounter{page}{1}

\appendices

\title{Fast and Informative Model Selection using Learning Curve Cross-Validation --- Supplemetary Material}
\IEEEtitleabstractindextext{}
{\maketitle}
\setcounter{figure}{0}

\section{Proof of Runtime Bound}\label{app:proof}
Here we show that the worst-case runtime of LCCV is at most twice as high as the one of a standard kCV under the following assumptions.
First, we assume that all considered anchor points are potentials of 2, and (here simplifying) that the full evaluation is also some such potential $2^d$.
Second, we assume that LCCV is configured to draw at most k samples at each anchor point.
Finally, we assume super-linearity of the runtime of the learners in the number of training instances.
Given the observations in \cite{mohr2021predicting}, this can be taken for granted for almost all learners, even though for very low runtimes, the overhead in the evaluation code could break this relationship in some cases.
To summarize, this means that the overall runtime of the validation function \valid for a learner is $\runtimecv = kg(2^d)$, where $g: \mathbb{N} \rightarrow \mathbb{R}$ is the runtime function of that learner.

Now let \runtimelccv be the time consumed by the LCCV algorithm.
For a number $c$ of considered anchors, a minimum exponent \minexp we want to use to learn, and $a_i$ being the \emph{number} of evaluations made at the anchor corresponding to exponent $i$, we have that
\begin{equation}
    \runtimelccv = \sum_{i=\minexp}^ca_ig(2^i) = g(1)\frac{k2^d}{k2^d} \sum_{i=\minexp}^ca_i2^i = kg(2^d)\sum_{i=\minexp}^c\frac{a_i}{k2^{d-i}}.
\end{equation}
For example, we could set $\minexp = 6$ to start with the anchor at 64 training instances.
The last term implies that the relationship between the two runtimes is
\begin{equation}
    \runtimelccv = \runtimecv \sum_{i=\minexp}^c\frac{a_i}{k2^{d-i}} \leq \runtimecv \sum_{i=0}^{d-\minexp}\frac{1}{2^i} < 2 \runtimecv,
\end{equation}
where the first inequality is based on the assumptions that
\begin{enumerate}
    \item $k$ is also a natural upper bound on the number of evaluations per anchor point and we hence have that $a_i \leq k$, and
    \item we run over all possible anchors.
\end{enumerate}
The last inequality simply follows from a comparison to the geometric series.

\section{Pipelines used in the Evaluation}
\label{sec:appendix:pipelines}
The pipelines we arm in this paper are from the search space used by the auto-sklearn tool \cite{feurer2015efficient}.

Almost every considered algorithm has hyperparameters, which are subject to optimization in our evaluation.
Since we cannot reasonably report the exact parameter setup even in the appendix, we refer the reader to the repository coming along with this paper in which all used parameters can be found in the \texttt{searchspace.json} JSON file.

\subsection{Considered Algorithms}
\label{sec:appendix:algorithms}
All applied algorithms are from the scikit-learn library \cite{pedregosa2011scikit}.

\medskip\noindent\textbf{data-pre-processors}: 
MinMaxScaler, Normalizer, PowerTransformer, QuantileTransformer, RobustScaler, StandardScaler, VarianceThreshold

\medskip\noindent\textbf{feature-pre-processors}: 
FastICA, FeatureAgglomeration, GenericUnivariateSelect, KernelPCA, Nystroem, PCA, PolynomialFeatures, RBFSampler, SelectPercentile

\medskip\noindent\textbf{classifiers}: 
BernoulliNB, DecisionTreeClassifier, ExtraTreesClassifier, GaussianNB, GradientBoostingClassifier, KNeighborsClassifier, LinearDiscriminantAnalysis, MLPClassifier, MultinomialNB, PassiveAggressiveClassifier, QuadraticDiscriminantAnalysis, RandomForestClassifier, SGDClassifier, SVC

\subsection{Pipeline Sampling}
\label{sec:pipelinesampling}
From the above pool of algorithms, the pipeline sequence is sampled i.i.d. where each candidate is sampled using the following routine:
\begin{enumerate}
    \item flip a fair coin to decide whether a data-pre-processor is used or not
    
    \item flip a fair coin to decide whether a feature-pre-processor is used or not
    
    \item for each slot of the pipeline that is supposed to be filled, sample one of the available algorithms based on their weight.
    The weight is computed based on the size of the hyperparameter-space of a component because we want to give more weight to algorithms with a big parameter space.
    The weight of an algorithm is the product of the weights of its parameters, which are
    \begin{itemize}
        \item the number of possible values for categorical hyperparameters and
        \item $\min\{10, s\}$ for a numeric parameter with $s$ possible values (typically $s = \infty$ but not for integer parameters with small domain).
    \end{itemize}
\end{enumerate}
We used the same 10 sequences on all datasets, and the sequences can be found at

\noindent
\url{https://github.com/fmohr/lccv/tree/master/publications/2022TPAMI/pipeline-sequences}

\section{Detailed Results of the Random Search}
\label{sec:appendix:results:randomsearch}
\begin{figure*}[ht!]
    \centering
    \includegraphics[width=\textwidth]{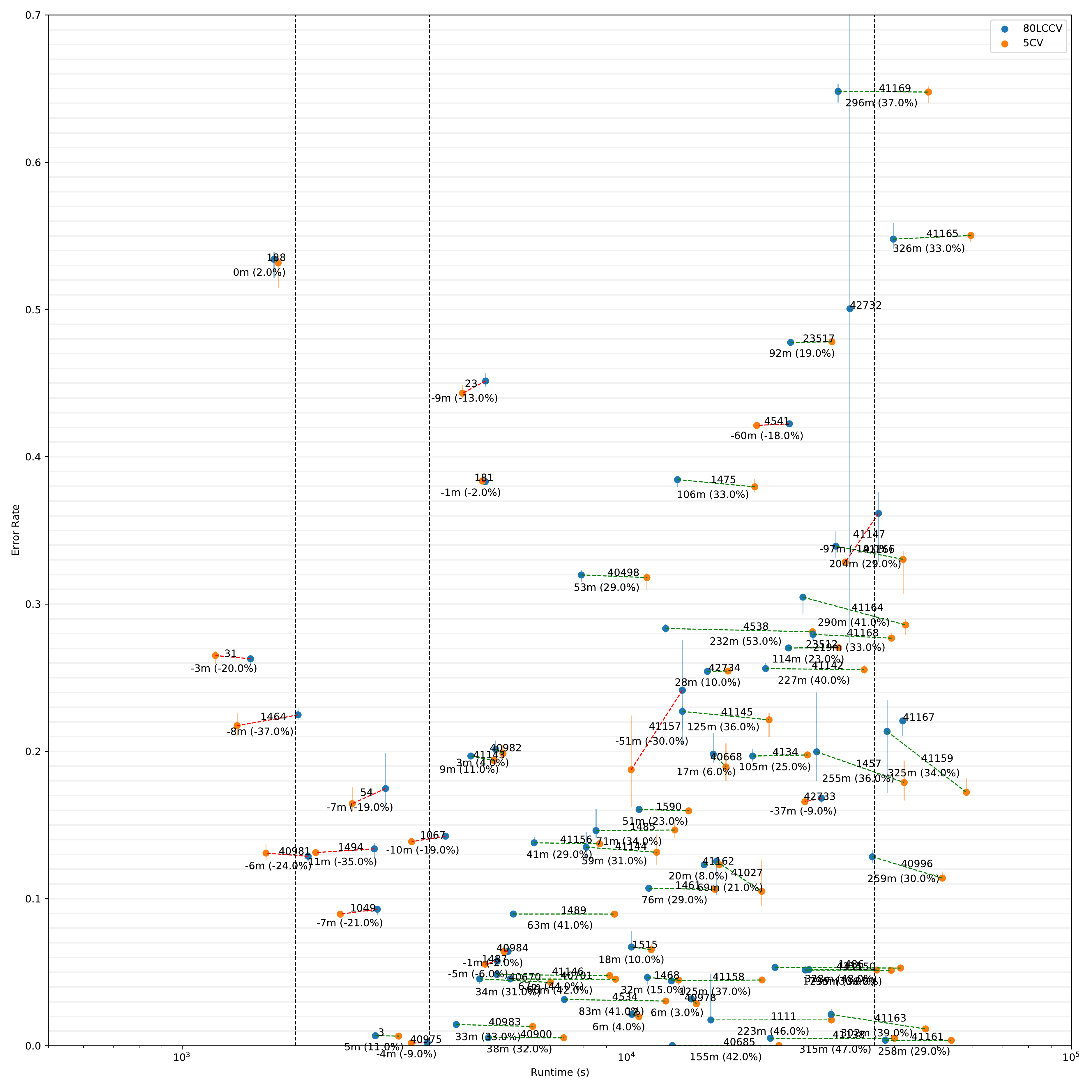}
    \caption{Visual comparison of random search using 80LCCV (blue) and using 5CV (orange) as validation technique. The x-axis displays the runtime in seconds (log-scale!), the y-axes displays the error rate. The dashed lines link the two observations for 80LCCV and 5CV that refer to the same dataset respectively (denoted at the edge label).
    Vertical colored lines connect the 10 and 90 percentiles.
    Vertical black lines are aids for 30m, 1h and 10h runtimes.}
    \label{fig:results-scatter:80}
\end{figure*}

\begin{figure*}[ht!]
    \centering
    \includegraphics[width=\textwidth]{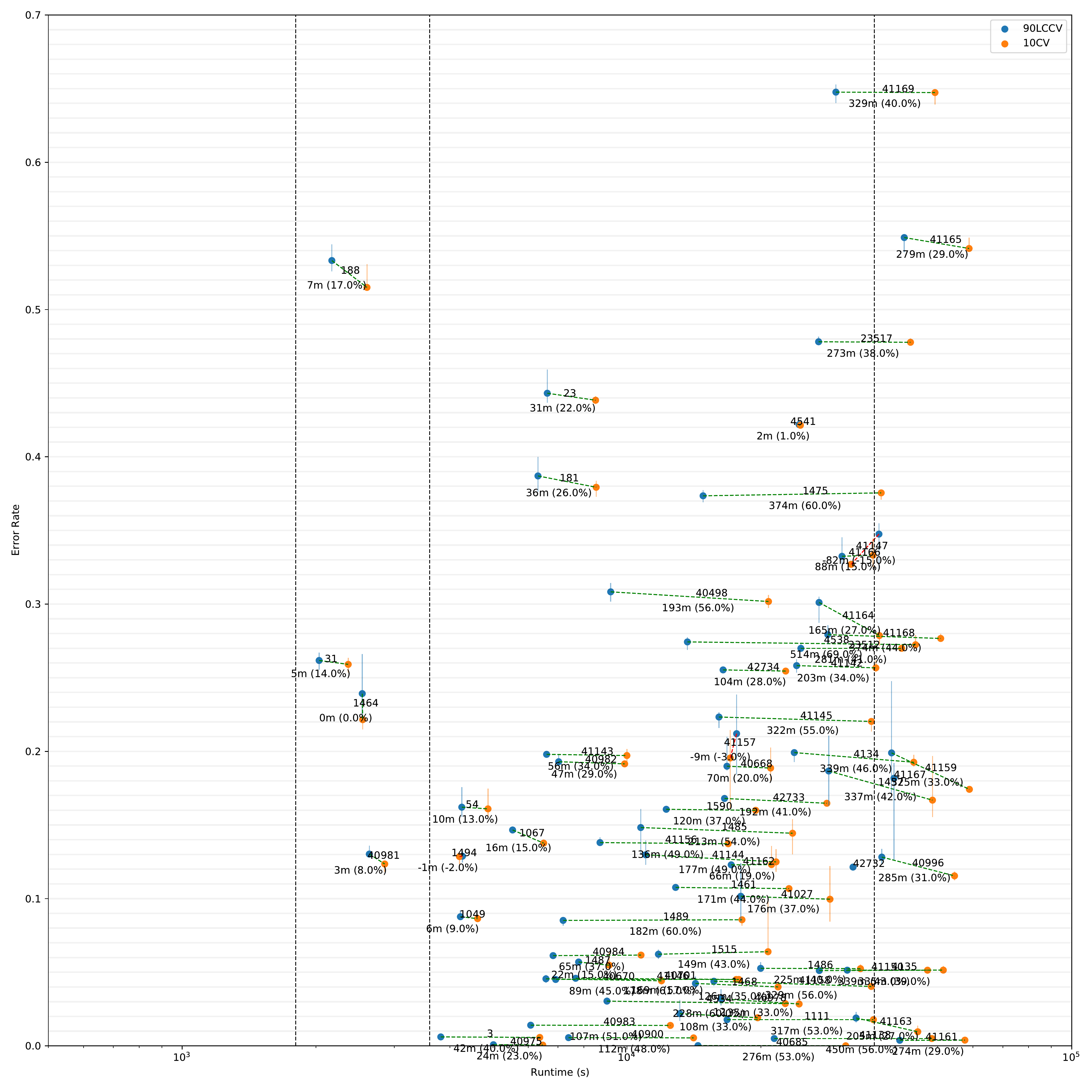}
    \caption{Visual comparison of random search using 90LCCV (blue) and using 10CV (orange) as validation technique. The x-axis displays the runtime in seconds (log-scale!), the y-axes displays the error rate. The dashed lines link the two observations for 90LCCV and 10CV that refer to the same dataset respectively (denoted at the edge label).
    Vertical colored lines connect the 10 and 90 percentiles.
    Vertical black lines are aids for 30m, 1h and 10h runtimes.
    }
    \label{fig:results-scatter:90}
\end{figure*}

The detailed results of the four validation methods per dataset are shown in Fig. \ref{fig:results-scatter:80} (5CV vs 80LCCV) and \ref{fig:results-scatter:90} (10CV vs 90LCCV), respectively.
The semantics of the figures is as follows.
Each pair of points connected by a dashed line stand for evaluations of a dataset.
In this pair, the blue point shows the mean runtime of the random search applying LCCV (x-axis) and the eventual mean error rate (y-axis).
The orange point shows the same information when validating with 10CV.
Note that the runtimes are shown on a log-scale.
The dashed lines connecting a pair of points are green if LCCV is faster than CV and red otherwise.
The line is labeled with the dataset id and the absolute (and relative) difference in runtime.
The \emph{vertical} black dashed lines are visual aids in the log-scale to mark runtimes of 30 minutes, 1 hour, and 10 hours.
Vertical colored lines connect the 10 and 90 percentiles for the results of the respective approach.

The same results are also given in Tables \ref{tab:results:random:80} (5CV vs 80LCCV) and \ref{tab:results:random:90} (10CV vs 90LCCV).
The table augments the information of the plot by listing the standard deviations for the different observations.
Additionally, we provide a reference for each of the analyzed datasets.

\section{Sensitivity Analysis of LCCV Parameters}
\label{sec:sensitivityanalysis}
\begin{figure*}[ht]
    \centering
    \includegraphics[width=.475\textwidth]{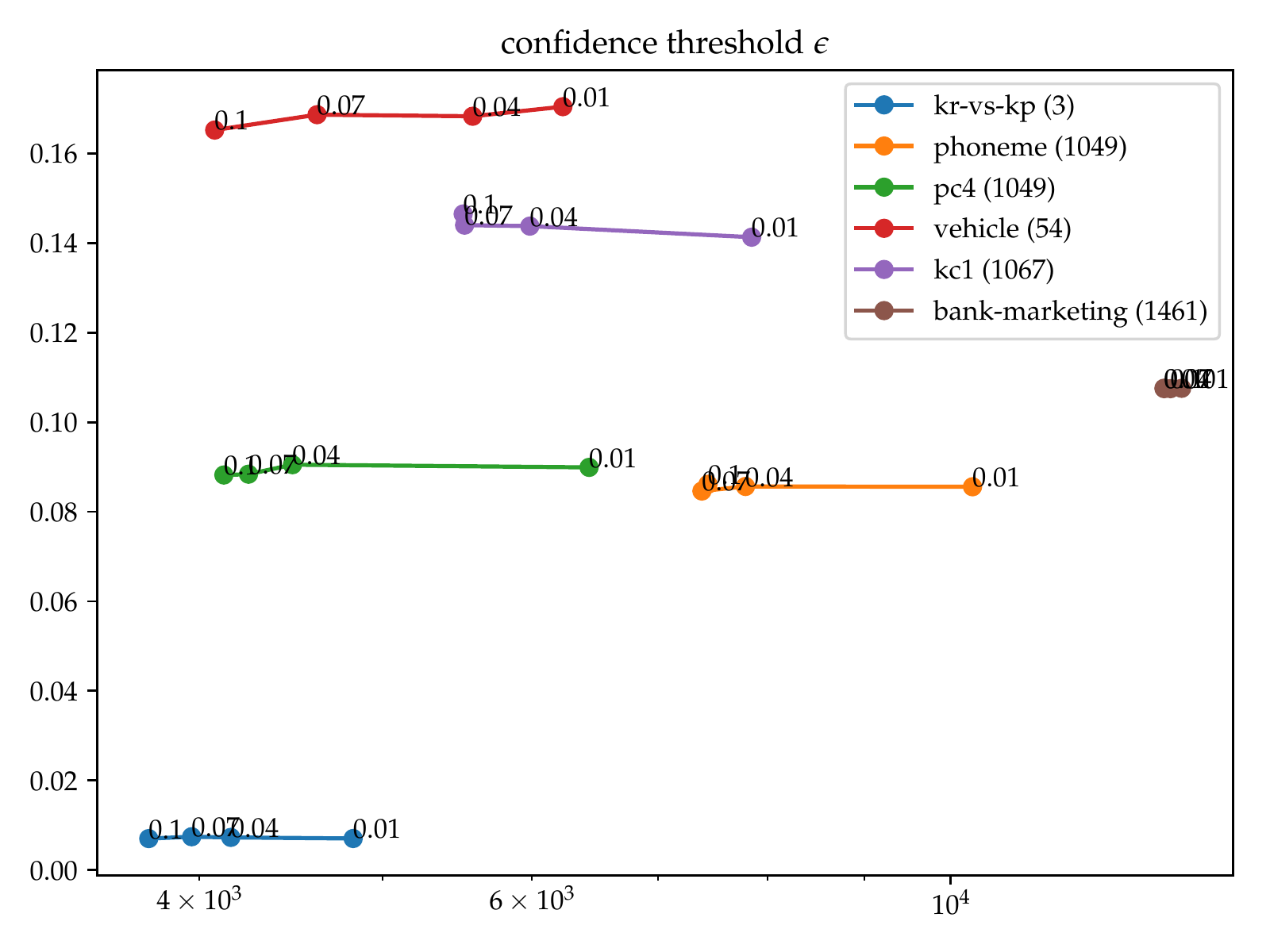}
    \includegraphics[width=.475\textwidth]{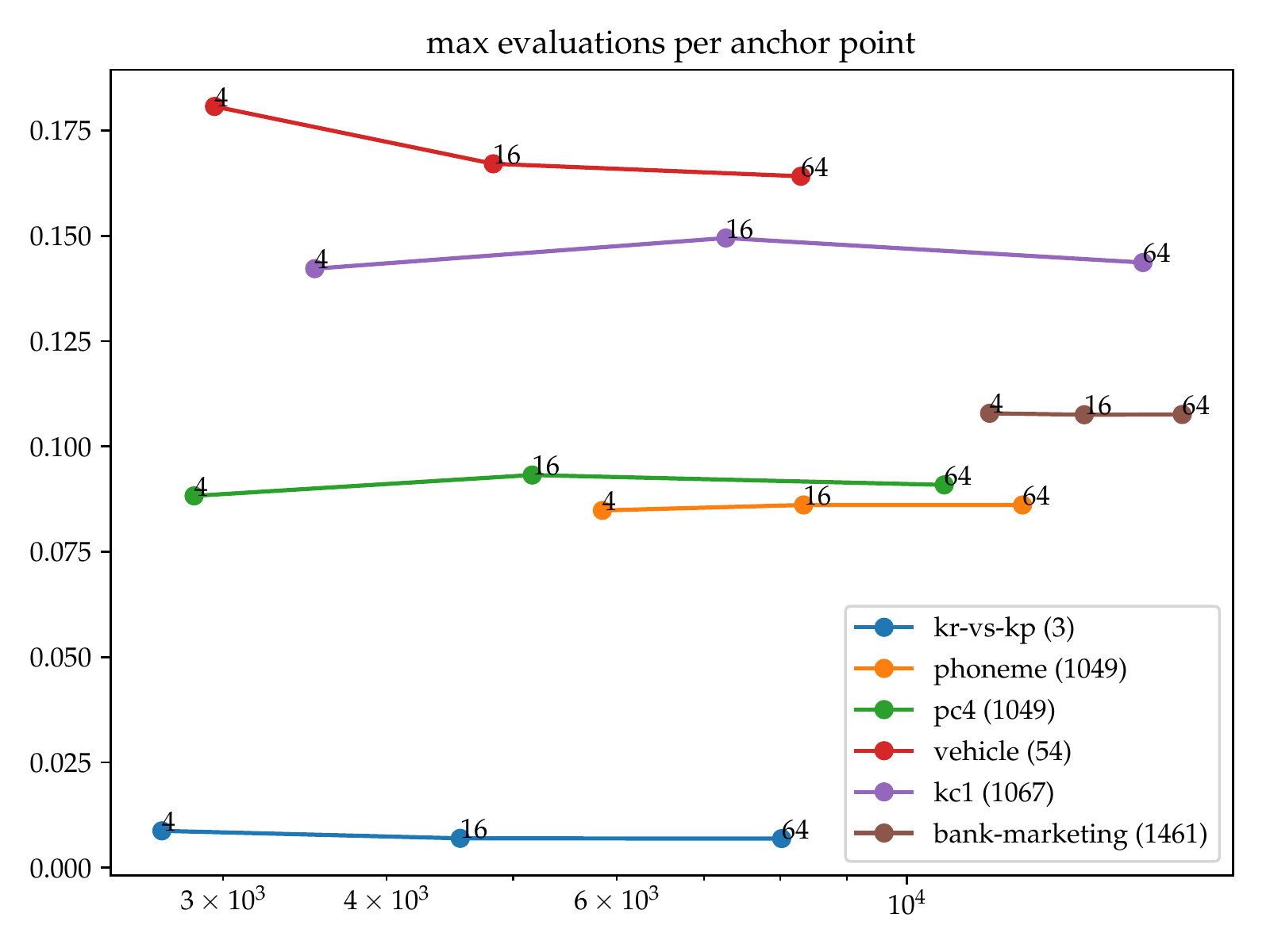}
    \includegraphics[width=.475\textwidth]{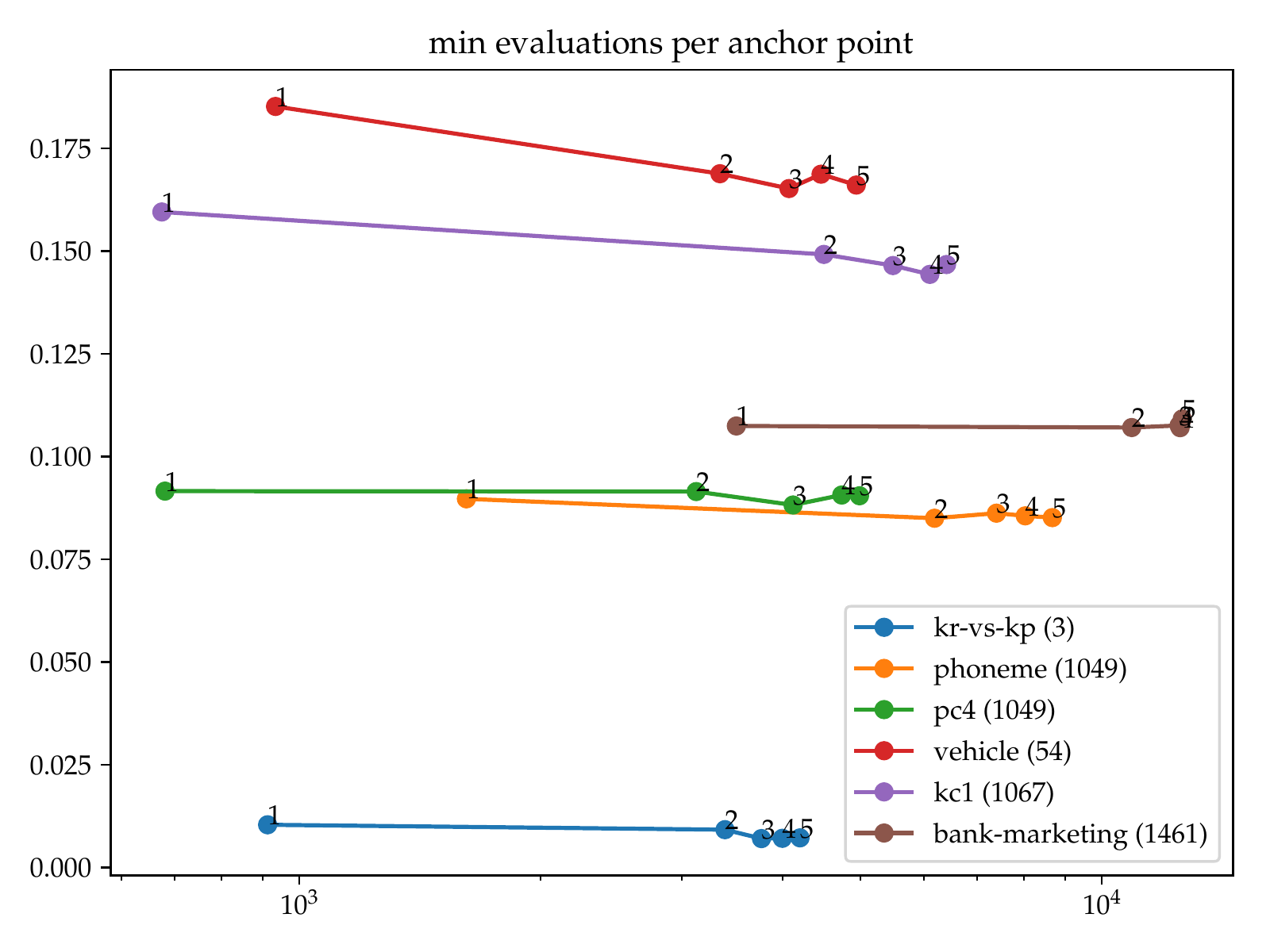}
    \includegraphics[width=.475\textwidth]{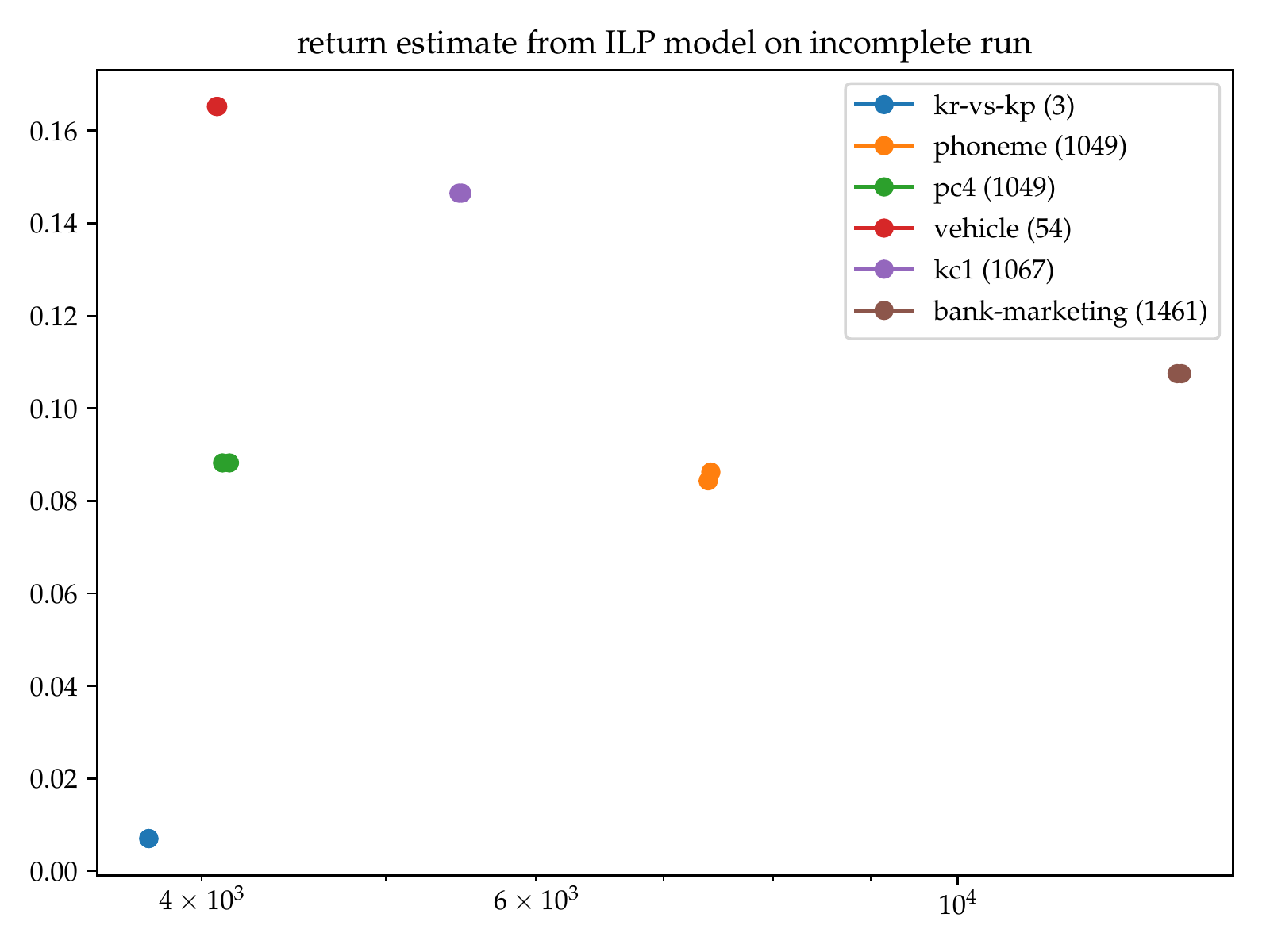}
    \caption{Sensitivity analysis of the hyperparameter of LCCV. Each dot represents a configuration of the hyperparameters of LCCV, averaged over 10 random seeds. \label{fig:sensitivity}}
\end{figure*}

We show the results of the sensitivity analysis.
We have selected several hyperparameters of LCCV and instantiated LCCV with several values for them. 
Each of those hyperparameters was considered in isolation; all other hyperparameters were kept constant.
The remaining setup was identical to the one in the main evaluation:
In each instantiation, LCCV evaluated 200 pipelines (the same pipelines as described in Appendix~\ref{sec:pipelinesampling}).
Each pipeline was allocated 300 seconds of run time. 
After the best pipeline has been determined, it is evaluated on MCCV to determine the final error rate. 
Figure~\ref{fig:sensitivity} shows the results.


Confidence threshold $\varepsilon$ controls the tolerated uncertainty at a certain anchor point. 
If the evaluations at a certain anchor point have confidence bounds higher than this threshold, more evaluations at this anchor point will be performed. 
Note that the number of evaluations at each anchor point is also bounded by a minimum and a maximum, which can be set by hyperparameters that are also visualized (second row). 
For these three hyperparameters, the results seem to be as expected.
Requiring higher confidence per anchor point or more evaluations results in a higher runtime. 
The error rate is hardly affected by this. 


Finally, we defined a hyperparameter that determined the policy of a certain configuration could not be finished due to run time restrictions.
In case of a time out, LCCV can of course always return the performance measured on the last successful anchor point. 
Alternatively, it can use the IPL-model to predict the final evaluation.
The analysis revealed that this hyperparameter has limited effect, 
but it should be noted that it was only evaluated on relatively small datasets. 
On larger datasets, the effect might be larger.

\section{Convexity of Learning Curves}\label{app:convexity}
In the following, we show the learning curves for all learners on all examined datasets.
Note that the y-scale is not always $[0,1]$ but is sometimes adjusted to make the behavior of the curves more visible.
The code for reproducing them can be found at \url{github.com/fmohr/lccv}.

The figures support the convexity claim for the large majority of cases.
92\% of the curves have a strictly convex behavior at \emph{any} observed points.
Among the rest, several curves are still convex in large parts, and only a tiny portion of curves show a ``heavily'' non-convex behavior.
None of these is even close to being competitive.

\clearpage
\newpage

\begin{table*}[ht!]
    \centering
    \footnotesize
    \input{results/results_80}
    \caption{Error rates and runtimes (s) of the Random Search with 5CV and 80LCCV.}
    \label{tab:results:random:80}
\end{table*}

\begin{table*}[ht!]
    \centering
    \footnotesize
    \input{results/results_90}
    \caption{Error rates and runtimes (s) of the Random Search with 10CV and 90LCCV.}
    \label{tab:results:random:90}
\end{table*}

\clearpage
\newpage

\clearpage
\newpage

\begin{center}
    \includegraphics[width=.95\textwidth]{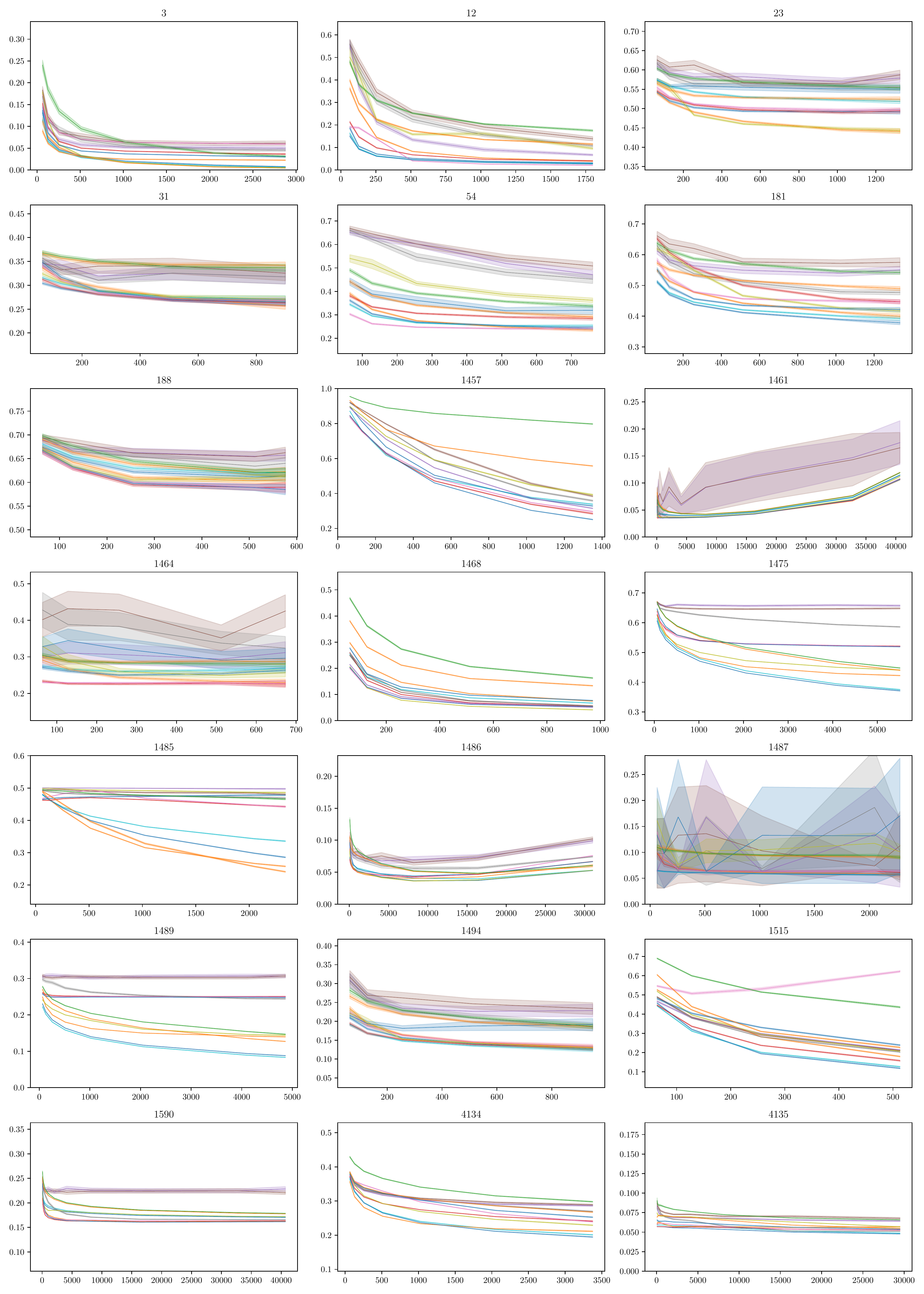}
\end{center}
\clearpage

\begin{center}
    \includegraphics[width=.95\textwidth]{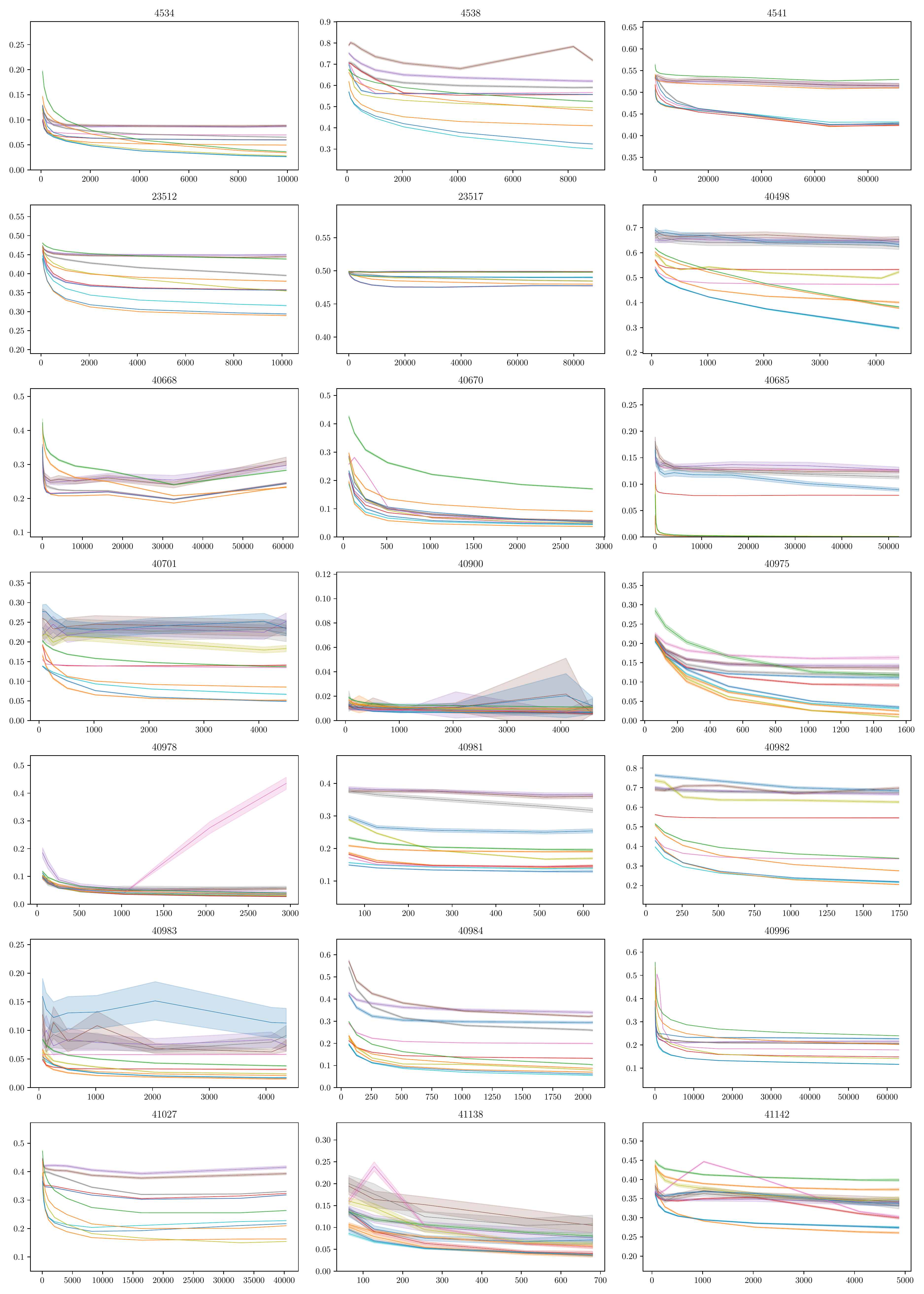}
\end{center}
\clearpage

\begin{center}
    \includegraphics[width=.95\textwidth]{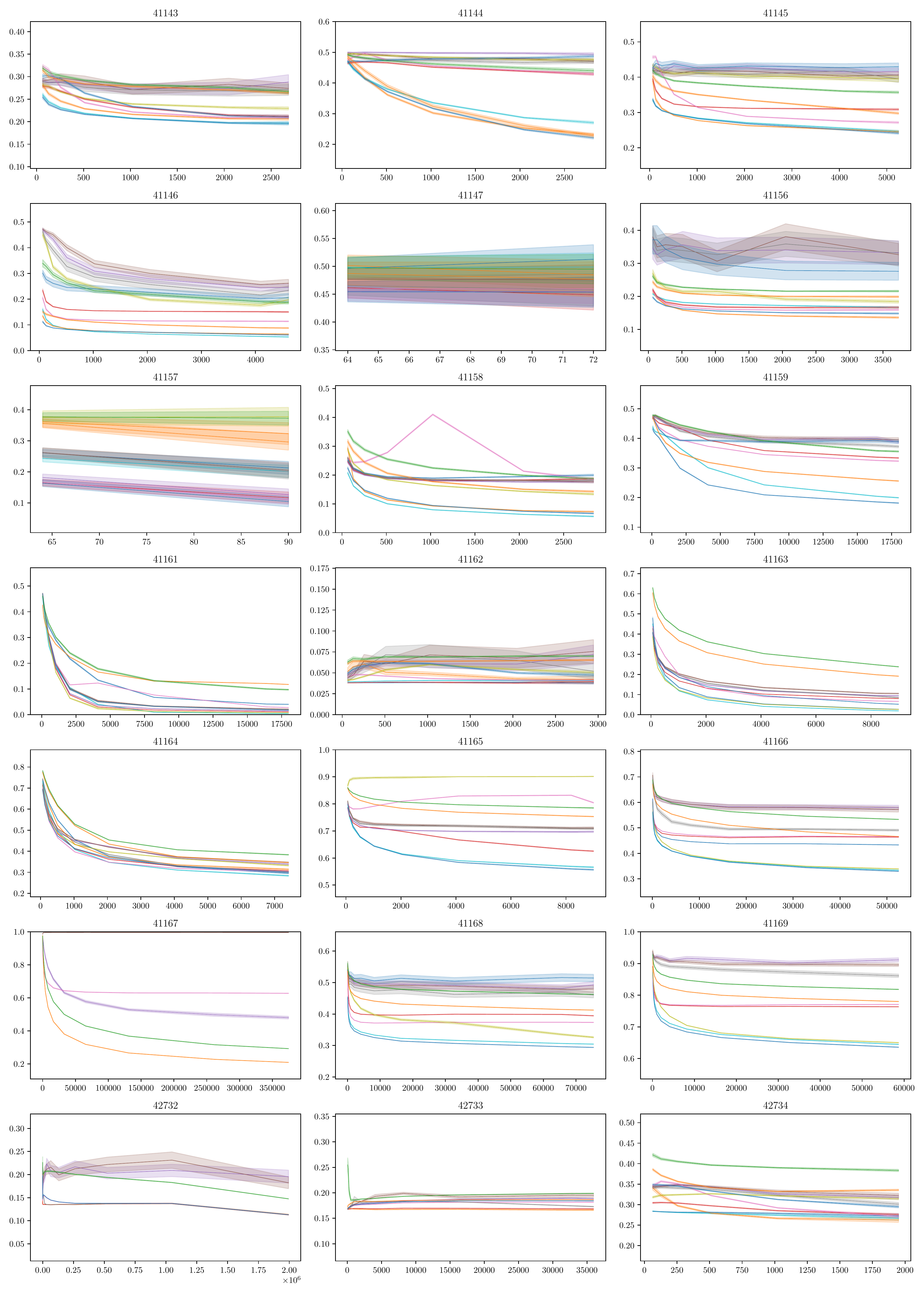}
\end{center}
\end{document}

%% file: lccv-algo.tex
\begin{algorithm}[t]
\scriptsize
  $(\anchor_1,..,\anchor_T) \leftarrow$ initialize anchor points from min\_exp and data size\;
  $(C_1,..,C_T) \leftarrow$ initialize confidence intervals as $[0,1]$ each\;
  $t \leftarrow 1$\;
  \While{$t \leq T \land (\sup{C_T} - \inf{C_T} > \maxciwidth) \land |O_T| < n$}{
      repair\_convexity $\leftarrow$ false\;
      
      \BlankLine
      \BlankLine
      \tcc{gather samples at current anchor point $\anchor_t$}
      \While{$\sup{C_t} - \inf{C_t} > \maxciwidth \land |O_t| < n \land \neg$ repair\_convexity}{\label{algline:stage:start}
        add sample for $s_t$ training points to $O_t$\;
        update confidence interval $C_t$\label{algline:confidencebounds}\;
        \lIf{$t > 1$}{
            $\sigma_{t-1} = (\sup{C_{t-1} - \inf{C_t}) / (\anchor_{t-1} - \anchor_t)}$
        }
        \If{$t > 2 \land \sigma_{t-1} < \sigma_{t-2} \land |O_{t-1}| < n$}{\label{algline:slopescan}
            repair\_convexity $\leftarrow$ true\;
        }
      }\label{algline:stage:end}
      
      \BlankLine
      \BlankLine
      \tcc{Decide how to proceed from this anchor}
      \uIf{repair\_convexity}{\label{algline:repair}
        $t \leftarrow t - 1$\;
        }
      \uElseIf{projected bound for $\anchor_T$ is $> \prunethreshold$}{\label{algline:prune}
        \Return \void
      }
         \uElseIf{$r = 1 \lor (t \geq 3~ \land \text{\sc ipl\_estimate}(\anchor_T) \leq \prunethreshold)$}{\label{algline:skip}
        $t \leftarrow T$\;
    }
    \Else{\label{algline:standardcase}
        $t \leftarrow t + 1$
      }
  }
  \Return $\mathit{\langle mean(C_T), (C_1,..,C_T) \rangle}$
  
  \caption{\footnotesize LCCV: LearningCurveCrossValidation}
  \label{algo:lccv}
\end{algorithm}

%% file: results/results_80.tex
\begin{tabular}{rrrrr}
\toprule
 openmlid & LCCV Performance &   LCCV Runtime & Baseline CV Perf & Baseline CV Runtime \\
\midrule
        3 &    0.01$\pm$0.00 &   2719$\pm$649 &    0.01$\pm$0.00 &        3068$\pm$765 \\
       12 &    0.02$\pm$0.00 &  10266$\pm$638 &    0.02$\pm$0.00 &      10639$\pm$1521 \\
       23 &    0.45$\pm$0.01 &   4812$\pm$744 &    0.44$\pm$0.00 &        4268$\pm$615 \\
       31 &    0.26$\pm$0.00 &   1425$\pm$432 &    0.26$\pm$0.01 &        1188$\pm$362 \\
       54 &    0.18$\pm$0.02 &   2867$\pm$513 &    0.17$\pm$0.01 &        2412$\pm$639 \\
      181 &    0.38$\pm$0.01 &   4805$\pm$750 &    0.38$\pm$0.00 &       4730$\pm$1205 \\
      188 &    0.53$\pm$0.01 &   1609$\pm$413 &    0.53$\pm$0.01 &        1644$\pm$559 \\
     1049 &    0.09$\pm$0.00 &   2746$\pm$489 &    0.09$\pm$0.00 &        2267$\pm$392 \\
     1067 &    0.14$\pm$0.00 &   3910$\pm$626 &    0.14$\pm$0.00 &        3280$\pm$675 \\
     1111 &    0.04$\pm$0.03 & 15443$\pm$1488 &    0.02$\pm$0.00 &      28824$\pm$6659 \\
     1457 &    0.21$\pm$0.03 & 26713$\pm$1798 &    0.18$\pm$0.02 &      42021$\pm$6654 \\
     1461 &    0.11$\pm$0.00 & 11199$\pm$1911 &    0.11$\pm$0.00 &      15791$\pm$4403 \\
     1464 &    0.23$\pm$0.01 &   1822$\pm$450 &    0.22$\pm$0.01 &        1329$\pm$394 \\
     1468 &    0.05$\pm$0.00 &  11119$\pm$643 &    0.04$\pm$0.00 &      13068$\pm$2284 \\
     1475 &    0.38$\pm$0.01 & 12992$\pm$1768 &    0.38$\pm$0.00 &      19384$\pm$3778 \\
     1485 &    0.15$\pm$0.01 &   8521$\pm$563 &    0.15$\pm$0.01 &      12818$\pm$1822 \\
     1486 &    0.05$\pm$0.00 & 21531$\pm$2946 &    0.05$\pm$0.00 &      41241$\pm$8726 \\
     1487 &    0.06$\pm$0.00 &   5109$\pm$658 &    0.06$\pm$0.00 &        4798$\pm$495 \\
     1489 &    0.09$\pm$0.00 &   5552$\pm$945 &    0.09$\pm$0.00 &       9377$\pm$2318 \\
     1494 &    0.13$\pm$0.01 &   2704$\pm$401 &    0.13$\pm$0.00 &        1996$\pm$290 \\
     1515 &    0.07$\pm$0.01 &  10234$\pm$932 &    0.07$\pm$0.01 &      11344$\pm$1960 \\
     1590 &    0.16$\pm$0.00 & 10652$\pm$2231 &    0.16$\pm$0.00 &      13772$\pm$2807 \\
     4134 &     0.2$\pm$0.00 & 19178$\pm$1337 &     0.2$\pm$0.00 &      25480$\pm$3971 \\
     4135 &    0.05$\pm$0.00 & 25669$\pm$5186 &    0.05$\pm$0.00 &      36459$\pm$6598 \\
     4534 &    0.03$\pm$0.00 &  7235$\pm$1456 &    0.03$\pm$0.00 &      12232$\pm$3274 \\
     4538 &    0.28$\pm$0.00 & 12220$\pm$1469 &    0.28$\pm$0.00 &      26149$\pm$4602 \\
     4541 &    0.42$\pm$0.00 & 23186$\pm$3544 &    0.42$\pm$0.00 &      19583$\pm$4706 \\
    23512 &    0.27$\pm$0.00 & 23085$\pm$2765 &    0.27$\pm$0.00 &      29928$\pm$5231 \\
    23517 &    0.48$\pm$0.00 & 23342$\pm$2841 &    0.48$\pm$0.00 &      28904$\pm$5655 \\
    40498 &    0.32$\pm$0.01 &   7896$\pm$862 &    0.33$\pm$0.05 &      11089$\pm$1530 \\
    40668 &     0.2$\pm$0.01 & 15640$\pm$2268 &    0.19$\pm$0.02 &      16700$\pm$2517 \\
    40670 &    0.05$\pm$0.00 &   4665$\pm$885 &    0.04$\pm$0.00 &       6735$\pm$1644 \\
    40685 &     0.0$\pm$0.00 & 12650$\pm$2009 &     0.0$\pm$0.00 &      21972$\pm$4829 \\
    40701 &    0.05$\pm$0.00 &   5462$\pm$949 &    0.04$\pm$0.00 &       9432$\pm$1563 \\
    40900 &    0.01$\pm$0.00 &   4879$\pm$921 &    0.01$\pm$0.00 &        7210$\pm$918 \\
    40975 &     0.0$\pm$0.00 &   3557$\pm$441 &     0.0$\pm$0.00 &        3275$\pm$602 \\
    40978 &    0.03$\pm$0.00 & 13958$\pm$1488 &    0.03$\pm$0.00 &      14331$\pm$3093 \\
    40981 &    0.13$\pm$0.01 &   1919$\pm$362 &    0.13$\pm$0.01 &        1543$\pm$471 \\
    40982 &     0.2$\pm$0.01 &   5068$\pm$547 &     0.2$\pm$0.01 &        5263$\pm$750 \\
    40983 &    0.01$\pm$0.00 &   4134$\pm$776 &    0.01$\pm$0.00 &        6140$\pm$947 \\
    40984 &    0.07$\pm$0.01 &   5402$\pm$772 &    0.06$\pm$0.00 &       5294$\pm$1232 \\
    40996 &    0.13$\pm$0.01 & 35647$\pm$1811 &    0.12$\pm$0.00 &      51214$\pm$8295 \\
    41027 &    0.12$\pm$0.02 & 15905$\pm$2470 &    0.11$\pm$0.02 &      20092$\pm$4308 \\
    41138 &    0.01$\pm$0.00 & 21020$\pm$4210 &    0.01$\pm$0.00 &      39950$\pm$8626 \\
    41142 &    0.26$\pm$0.01 & 20504$\pm$1226 &    0.26$\pm$0.01 &      34163$\pm$5912 \\
    41143 &     0.2$\pm$0.00 &   4457$\pm$779 &    0.19$\pm$0.00 &        5026$\pm$957 \\
    41144 &    0.13$\pm$0.01 &   8097$\pm$916 &    0.13$\pm$0.01 &      11666$\pm$1701 \\
    41145 &    0.22$\pm$0.01 & 13328$\pm$1060 &    0.22$\pm$0.01 &      20875$\pm$2270 \\
    41146 &    0.05$\pm$0.00 &   5097$\pm$702 &    0.05$\pm$0.00 &       9146$\pm$1898 \\
    41147 &    0.36$\pm$0.03 & 36794$\pm$3679 &    0.33$\pm$0.00 &         30966$\pm$0 \\
    41150 &    0.05$\pm$0.00 & 25190$\pm$2217 &    0.05$\pm$0.00 &      39296$\pm$5996 \\
    41156 &    0.14$\pm$0.00 &   6186$\pm$732 &    0.14$\pm$0.00 &       8682$\pm$1708 \\
    41157 &    0.24$\pm$0.04 & 13326$\pm$1148 &    0.19$\pm$0.03 &      10220$\pm$1658 \\
    41158 &    0.05$\pm$0.01 &  12591$\pm$835 &    0.04$\pm$0.00 &      20127$\pm$4137 \\
    41159 &    0.21$\pm$0.03 & 38455$\pm$2073 &    0.18$\pm$0.01 &     57982$\pm$10141 \\
    41161 &     0.0$\pm$0.00 & 38107$\pm$2105 &     0.0$\pm$0.00 &      53635$\pm$3355 \\
    41162 &    0.12$\pm$0.00 & 14920$\pm$1352 &    0.13$\pm$0.03 &      16137$\pm$2448 \\
    41163 &    0.02$\pm$0.01 & 28787$\pm$1586 &    0.03$\pm$0.04 &     46928$\pm$10320 \\
    41164 &     0.3$\pm$0.01 & 24875$\pm$2143 &    0.29$\pm$0.01 &     42334$\pm$11041 \\
    41165 &    0.55$\pm$0.01 & 39723$\pm$2033 &    0.55$\pm$0.00 &      59340$\pm$7449 \\
    41166 &    0.34$\pm$0.01 & 29502$\pm$2300 &    0.32$\pm$0.02 &      41780$\pm$9741 \\
    41167 &    0.22$\pm$0.01 & 41713$\pm$1992 &              NaN &               86400 \\
    41168 &    0.28$\pm$0.02 & 26216$\pm$2170 &    0.28$\pm$0.00 &      39360$\pm$5874 \\
    41169 &    0.65$\pm$0.01 & 29847$\pm$2423 &    0.65$\pm$0.01 &      47628$\pm$8652 \\
    42732 &     0.5$\pm$0.38 & 31728$\pm$1350 &              NaN &               86400 \\
    42733 &    0.17$\pm$0.00 & 27371$\pm$5504 &    0.17$\pm$0.00 &      25123$\pm$6305 \\
    42734 &    0.25$\pm$0.00 & 15167$\pm$2144 &    0.25$\pm$0.00 &      16884$\pm$2830 \\
\bottomrule
\end{tabular}

%% file: results/results_90.tex
\begin{tabular}{rrrrr}
\toprule
 openmlid & LCCV Performance &   LCCV Runtime & Baseline CV Perf & Baseline CV Runtime \\
\midrule
        3 &    0.01$\pm$0.00 &   3814$\pm$801 &    0.01$\pm$0.00 &       6368$\pm$2295 \\
       12 &    0.02$\pm$0.01 &  13150$\pm$649 &    0.02$\pm$0.00 &      19676$\pm$4979 \\
       23 &    0.45$\pm$0.01 &   6620$\pm$805 &    0.44$\pm$0.01 &       8502$\pm$1981 \\
       31 &    0.26$\pm$0.01 &   2032$\pm$421 &    0.26$\pm$0.01 &        2362$\pm$736 \\
       54 &    0.17$\pm$0.01 &   4253$\pm$823 &    0.17$\pm$0.01 &       4871$\pm$1687 \\
      181 &    0.39$\pm$0.02 &   6309$\pm$792 &    0.38$\pm$0.01 &       8528$\pm$1873 \\
      188 &    0.54$\pm$0.01 &   2170$\pm$526 &    0.52$\pm$0.01 &       2603$\pm$1162 \\
     1049 &    0.09$\pm$0.00 &   4222$\pm$706 &    0.09$\pm$0.00 &        4615$\pm$924 \\
     1067 &    0.15$\pm$0.00 &   5532$\pm$636 &    0.14$\pm$0.00 &       6498$\pm$1454 \\
     1111 &    0.02$\pm$0.00 & 16796$\pm$3942 &    0.02$\pm$0.00 &     35830$\pm$13662 \\
     1457 &    0.19$\pm$0.03 & 28425$\pm$1892 &    0.17$\pm$0.02 &     48648$\pm$14121 \\
     1461 &    0.11$\pm$0.00 & 12866$\pm$1751 &    0.11$\pm$0.00 &     23141$\pm$11020 \\
     1464 &    0.24$\pm$0.02 &   2540$\pm$555 &    0.22$\pm$0.01 &        2546$\pm$734 \\
     1468 &    0.04$\pm$0.01 &  14258$\pm$956 &    0.04$\pm$0.00 &      21870$\pm$7024 \\
     1475 &    0.37$\pm$0.00 & 14834$\pm$1741 &    0.38$\pm$0.00 &     37304$\pm$10756 \\
     1485 &    0.15$\pm$0.02 &  10740$\pm$826 &    0.14$\pm$0.01 &      23574$\pm$6048 \\
     1486 &    0.05$\pm$0.00 & 19985$\pm$3222 &    0.05$\pm$0.00 &      33505$\pm$9037 \\
     1487 &    0.06$\pm$0.00 &  7786$\pm$1329 &    0.06$\pm$0.00 &       9123$\pm$1213 \\
     1489 &    0.08$\pm$0.00 &   7187$\pm$944 &    0.08$\pm$0.00 &      18143$\pm$6255 \\
     1494 &    0.13$\pm$0.01 &   4267$\pm$529 &    0.13$\pm$0.01 &        4202$\pm$890 \\
     1515 &    0.06$\pm$0.01 & 11769$\pm$1083 &    0.08$\pm$0.02 &      20764$\pm$6681 \\
     1590 &    0.16$\pm$0.00 & 12257$\pm$2578 &    0.16$\pm$0.00 &      19477$\pm$7185 \\
     4134 &     0.2$\pm$0.01 & 23783$\pm$1781 &    0.19$\pm$0.00 &     44151$\pm$13991 \\
     4135 &    0.05$\pm$0.00 & 31290$\pm$5924 &    0.05$\pm$0.00 &     51449$\pm$14106 \\
     4534 &    0.03$\pm$0.00 &  9018$\pm$1721 &    0.03$\pm$0.00 &     22702$\pm$10760 \\
     4538 &    0.27$\pm$0.01 & 13676$\pm$1654 &    0.27$\pm$0.00 &     44568$\pm$18258 \\
     4541 &    0.42$\pm$0.00 & 24387$\pm$3276 &    0.42$\pm$0.00 &      24549$\pm$7060 \\
    23512 &    0.27$\pm$0.00 & 24611$\pm$3028 &    0.27$\pm$0.00 &     41472$\pm$13729 \\
    23517 &    0.48$\pm$0.00 & 26976$\pm$3174 &    0.48$\pm$0.00 &     43414$\pm$12728 \\
    40498 &    0.31$\pm$0.01 &  9194$\pm$1025 &     0.3$\pm$0.01 &      20812$\pm$4968 \\
    40668 &     0.2$\pm$0.02 & 16783$\pm$2447 &    0.19$\pm$0.02 &      21041$\pm$4837 \\
    40670 &    0.05$\pm$0.01 &  6575$\pm$1120 &    0.04$\pm$0.00 &      11958$\pm$2691 \\
    40685 &    0.01$\pm$0.02 & 14447$\pm$2132 &     0.0$\pm$0.00 &      31041$\pm$8996 \\
    40701 &    0.05$\pm$0.00 &  7668$\pm$1448 &    0.05$\pm$0.00 &      17849$\pm$4633 \\
    40900 &    0.01$\pm$0.00 &  7391$\pm$1098 &    0.01$\pm$0.00 &      14121$\pm$2842 \\
    40975 &     0.0$\pm$0.00 &   5009$\pm$623 &     0.0$\pm$0.00 &       6467$\pm$1378 \\
    40978 &    0.03$\pm$0.00 & 16274$\pm$1429 &    0.03$\pm$0.00 &      24388$\pm$8456 \\
    40981 &    0.13$\pm$0.01 &   2635$\pm$499 &    0.12$\pm$0.00 &        2854$\pm$525 \\
    40982 &    0.19$\pm$0.01 &   7022$\pm$966 &     0.2$\pm$0.01 &       9883$\pm$1591 \\
    40983 &    0.01$\pm$0.00 &  6075$\pm$1156 &    0.01$\pm$0.00 &      12522$\pm$3087 \\
    40984 &    0.06$\pm$0.00 &   6823$\pm$898 &    0.06$\pm$0.01 &      10757$\pm$2836 \\
    40996 &    0.13$\pm$0.01 & 37418$\pm$1205 &    0.12$\pm$0.00 &      54525$\pm$8923 \\
    41027 &     0.1$\pm$0.02 & 18029$\pm$2216 &    0.11$\pm$0.04 &      28614$\pm$7346 \\
    41138 &    0.01$\pm$0.00 & 21440$\pm$4014 &    0.01$\pm$0.00 &     48473$\pm$16425 \\
    41142 &    0.26$\pm$0.01 & 24077$\pm$1628 &    0.26$\pm$0.00 &      36286$\pm$2687 \\
    41143 &     0.2$\pm$0.00 &  6597$\pm$1193 &     0.2$\pm$0.01 &       9998$\pm$2617 \\
    41144 &    0.13$\pm$0.01 & 11018$\pm$1508 &    0.13$\pm$0.01 &      21647$\pm$5049 \\
    41145 &    0.22$\pm$0.01 & 16103$\pm$1226 &    0.22$\pm$0.01 &      35455$\pm$8022 \\
    41146 &    0.04$\pm$0.00 &  6915$\pm$1145 &    0.04$\pm$0.00 &      17615$\pm$5001 \\
    41147 &    0.34$\pm$0.02 & 36872$\pm$1991 &    0.33$\pm$0.00 &         31924$\pm$0 \\
    41150 &    0.05$\pm$0.00 & 27072$\pm$2627 &    0.05$\pm$0.00 &     47412$\pm$12695 \\
    41156 &    0.14$\pm$0.00 &  8699$\pm$1113 &    0.14$\pm$0.00 &      16892$\pm$5729 \\
    41157 &    0.21$\pm$0.03 & 17644$\pm$1717 &    0.19$\pm$0.03 &      17073$\pm$3532 \\
    41158 &    0.04$\pm$0.00 &  15687$\pm$898 &    0.04$\pm$0.00 &     35441$\pm$11638 \\
    41159 &    0.21$\pm$0.04 & 39357$\pm$2688 &    0.17$\pm$0.00 &      58890$\pm$4620 \\
    41161 &     0.0$\pm$0.00 & 41060$\pm$1521 &     0.0$\pm$0.00 &      57535$\pm$5450 \\
    41162 &    0.12$\pm$0.00 & 17158$\pm$1553 &    0.14$\pm$0.04 &      21137$\pm$4880 \\
    41163 &    0.02$\pm$0.00 & 32752$\pm$2693 &    0.01$\pm$0.00 &      45077$\pm$2106 \\
    41164 &     0.3$\pm$0.01 & 27038$\pm$1511 &    0.28$\pm$0.01 &      36959$\pm$3353 \\
    41165 &    0.55$\pm$0.01 & 42028$\pm$1620 &    0.54$\pm$0.01 &      58794$\pm$3410 \\
    41166 &    0.33$\pm$0.01 & 30450$\pm$1840 &    0.33$\pm$0.01 &      35753$\pm$2358 \\
    41167 &    0.16$\pm$0.05 & 39846$\pm$1962 &              NaN &               86400 \\
    41168 &    0.28$\pm$0.01 & 28294$\pm$2250 &    0.28$\pm$0.00 &     50759$\pm$13764 \\
    41169 &    0.65$\pm$0.01 & 29502$\pm$2115 &    0.64$\pm$0.01 &     49281$\pm$12972 \\
    42732 &    0.12$\pm$0.00 &    32240$\pm$0 &              NaN &               86400 \\
    42733 &    0.17$\pm$0.00 & 16574$\pm$3090 &    0.17$\pm$0.00 &      28152$\pm$9530 \\
    42734 &    0.26$\pm$0.00 & 16452$\pm$1873 &    0.25$\pm$0.00 &      22749$\pm$6145 \\
\bottomrule
\end{tabular}